\pdfoutput=1

\documentclass[10pt,twocolumn,letterpaper]{article}

\usepackage[pagenumbers]{cvpr}

\usepackage{multirow}
\usepackage{makecell}
\usepackage{cuted}
\usepackage{afterpage}

\definecolor{cvprblue}{rgb}{0.21,0.49,0.74}
\usepackage[pagebackref,breaklinks,colorlinks,allcolors=cvprblue]{hyperref}

\title{RadarOSV: Unified 3D Scene and Motion Perception from mmWave Radars}

\author{
Yuan Fang$^{1}$ \quad Long Zhuang$^{2,3}$ \quad Xijia Wei$^{1}$ \quad Jingqi Jiang$^{4}$\\
Fangzhan Shi$^{1}$ \quad Simon Julier$^{1}$ \quad Kevin Chetty$^{1}$
}

\begin{document}
\maketitle
{\let\thefootnote\relax
\makeatletter\renewcommand{\@makefntext}[1]{\noindent #1}\makeatother
\footnotetext{$^{1}$University College London, London, United Kingdom.}
\footnotetext{$^{2}$Southeast University, Nanjing, China.}
\footnotetext{$^{3}$Nanyang Technological University, Singapore.}
\footnotetext{$^{4}$Imperial College London, London, United Kingdom.}
\footnotetext{Y.~Fang and L.~Zhuang contributed equally.}}

\begin{strip}
    \centering
    \includegraphics[width=\textwidth]{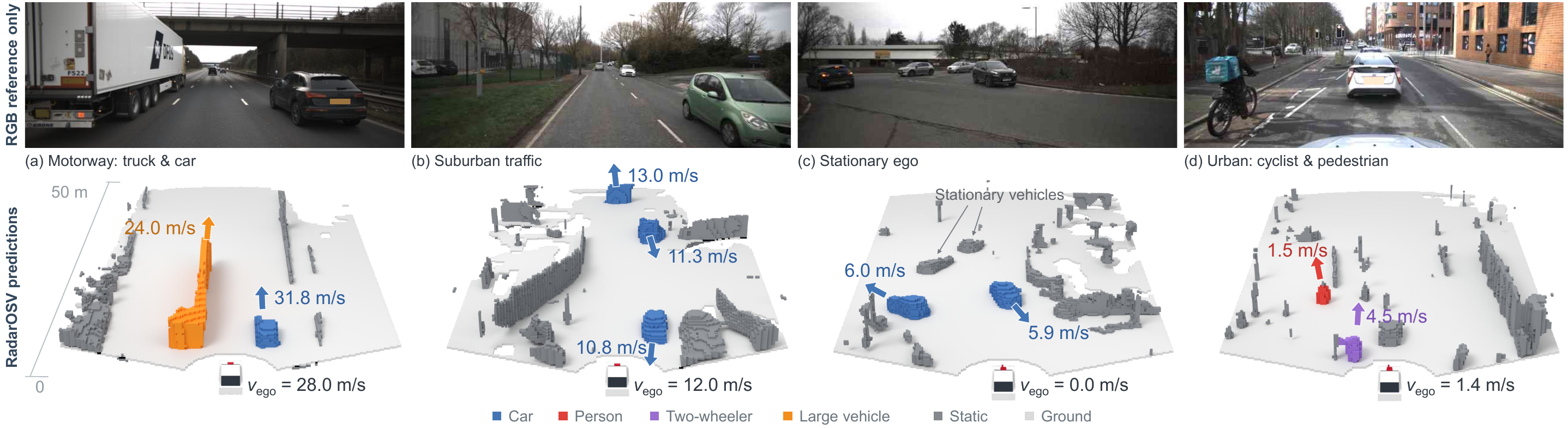}

    \captionsetup{hypcap=false}
    \captionof{figure}{
        \textbf{RadarOSV predictions in four driving scenes.}
        RadarOSV integrates complementary information from cascaded
        imaging radar and single-chip radar to jointly predict
        3D occupancy, dynamic-object semantics and velocities,
        as well as ego speed ($v_{\mathrm{ego}}$).
        Arrows show the moving direction, and each labeled speed is the
        median predicted velocity within the annotated object region.
        In (a)--(c), each connected component is colored by its majority class.
        Each panel covers the region up to 50\,m ahead.
        RGB images are for visual reference only.
    }
    \label{fig:radarosv_teaser}
    \captionsetup{hypcap=true}
\end{strip}

\begin{abstract}
Millimeter-wave radar offers weather-resilient sensing for 3D occupancy prediction. Existing radar occupancy methods mainly model geometry, with limited joint modeling of semantics and motion. Cascaded imaging radar provides fine spatial detail but has a limited unambiguous velocity range, whereas single-chip radar offers a wider range but coarser spatial resolution. We propose RadarOSV, to our knowledge the first radar-only framework that jointly predicts occupancy, dynamic object semantics, and ground-relative velocity from raw radar data in unified voxels, by exploiting complementary measurements from both radars. Its Polar-to-Cartesian Geometry (PCG) backbone maps polar imaging-radar features into Cartesian voxels through fixed and deformable sampling, guiding cross-radar fusion for geometry and attribute prediction. Single-chip detection outputs provide class evidence, while raw Doppler spectra supply motion features and radial-velocity candidates. The velocity readout selects from these candidates using fused features to resolve aliasing, then combines the selected radial estimate with learned motion. Ego-motion compensation uses radar-estimated velocity, so inference requires no external odometry or other sensors. On K-Radar, single-frame PCG achieves 40.7\% mIoU within 12.8~m, compared with RadarOcc's reported 34.3\%. On the BRIDGE-Liv dataset, compared with a cascaded-radar baseline predicting the same outputs with external ego velocity, RadarOSV improves macro-averaged accuracy across four dynamic object classes within annotated object regions from 63.1\% to 75.5\% and reduces object radial-velocity MAE against manual Doppler annotations from 1.03 to 0.51~m/s, while maintaining geometric occupancy mIoU.
\end{abstract}

\section{Introduction}
\label{sec:introduction}

Safe autonomous driving depends on accurate environmental perception and reliable identification of road users' categories and motion states. Camera and LiDAR measurements are commonly represented in bird's-eye view (BEV) for object detection and road-user perception~\cite{bevfusion}. However, bounding boxes and planar semantic maps cannot fully capture irregular obstacles. Vehicles must still determine whether these obstacles occupy the driving space, even when their categories are unknown. Voxel-based 3D occupancy prediction describes scene geometry and free space, and can incorporate semantic and motion information to support risk assessment and planning~\cite{surroundocc,viewformer}.

Existing occupancy prediction methods primarily rely on cameras and LiDAR, whose measurement quality can degrade in rain and fog, and cameras are further affected by low light and glare~\cite{seeing_through_fog}. Millimeter-wave radar operates independently of ambient illumination, is relatively resilient to rain and fog, and directly measures relative radial velocity through Doppler. Conventional automotive radar has mainly supported driver assistance such as adaptive cruise control and collision warning~\cite{ti_automotive_radar}. Advances in integrated transceivers and cascaded imaging have improved azimuth and elevation resolution, providing a hardware basis for 3D scene perception. Compared with sparse point clouds, radar tensors before point-cloud generation can preserve weak returns and scene details that detection filtering may discard. Related studies have shown their potential for 3D reconstruction and occupancy prediction~\cite{radelft,autolabel,radarocc}. However, jointly predicting scene occupancy, multi-class dynamic-object semantics, and velocity in unified 3D voxels using only radar at inference still requires further investigation.

Cascaded imaging radar (CAS) offers high azimuth and elevation resolution for detailed 3D geometric reconstruction. However, in common time-division-multiplexed cascaded configurations, sequential transmission from multiple transmit antennas increases the effective sampling interval, limiting the unambiguous velocity range~\cite{radelft}. Velocities beyond this range alias, making object motion harder to recover~\cite{radarocc}. In contrast, single-chip radar (SC) configurations with fewer transmit channels and shorter sampling intervals can provide a wider unambiguous velocity range. Their echo structure and Doppler distributions offer complementary cues for dynamic-object semantics and motion, but their lower spatial resolution limits the recovery of fine 3D structure~\cite{ramod}. The two radar types therefore provide complementary evidence for scene geometry and dynamic-object perception. Jointly exploiting these measurements requires associating class and motion evidence at different resolutions with common voxels, and using scene context to determine whether local Doppler returns correspond to object motion. Moreover, Doppler constrains only relative radial motion; recovering planar object velocity still requires learned motion estimates and ego-motion compensation.

We propose \textbf{RadarOSV}, a unified 3D scene and motion perception framework that takes raw CAS and SC tensors and jointly predicts scene \textbf{O}ccupancy, multi-class dynamic-object \textbf{S}emantics, and ground-relative \textbf{V}elocity in common voxels. The Polar-to-Cartesian Geometry (PCG) backbone encodes imaging-radar measurements in polar space and combines raw power evidence, fine-scale structure, and local context through fixed and learnable sampling to build a 3D geometric representation. PCG can independently perform imaging-radar occupancy prediction and provides spatial context for cross-radar association and joint prediction. Geometry-Guided Radar Fusion (GRF) uses this geometric context to aggregate single-chip features for 3D geometry decoding and object attribute prediction. Dynamic Semantic Readout (DSR) introduces object-level class cues from the single-chip branch into voxel classification, while Explicit Doppler Readout (EDR) supplies spectral features and radial-velocity candidates. The final velocity readout uses fused scene features to select candidates and handle velocity aliasing, and combines learned motion components with ego-motion compensation to recover ground-relative planar velocity. Ego velocity is also estimated from radar and refined using static returns, so the full system needs no cameras, LiDAR, or external odometry at inference.

We evaluate PCG's independent geometry modeling capability on K-Radar and RaDelft using imaging radar alone. Under the corresponding K-Radar evaluation settings, PCG exceeds the mIoU reported in the RadarOcc paper. On the dual-radar BRIDGE-Liv dataset, we compare single-radar and dual-radar configurations with the same task outputs and a common external ego-motion input, and conduct module ablations to examine the roles of complementary measurements and individual design choices. Incorporating single-chip measurements improves dynamic-object classification and radial-velocity estimation while preserving geometric reconstruction performance; the full system uses radar-based ego-motion estimation for radar-only inference.

Our main contributions are as follows:
\begin{enumerate}
    \item We propose RadarOSV for unified radar scene and motion perception. PCG builds a 3D representation for geometric reconstruction and cross-radar association, to jointly predict occupancy, multi-class dynamic-object semantics, and velocity in common voxels.
    \item We design GRF and DSR to aggregate cross-radar features using 3D geometric context and introduce object-level class evidence into voxel semantic prediction, supporting joint geometry and attribute prediction.
    \item We design EDR to provide spectral features and radial-velocity candidates. The final velocity readout uses scene features to select candidates and handle velocity aliasing, and combines learned motion estimates with radar-based ego-motion compensation to estimate ground-relative planar velocity using only radar.
\end{enumerate}

\section{Related Work}
\label{sec:related_work}

\begin{table*}[t]
  \caption{Inference inputs and output capabilities of radar-tensor-based 3D occupancy methods.}
  \label{tab:capability}
  \centering
  \scriptsize
  \setlength{\tabcolsep}{3pt}
  \setlength{\aboverulesep}{0.3pt}
  \setlength{\belowrulesep}{0.3pt}
  \renewcommand{\arraystretch}{1.1}

  \begin{tabular*}{\textwidth}{@{\extracolsep{\fill}}lllllcc@{}}
    \toprule
    Method & Radar input & External input &
    Occupancy classes & Object classes &
    \makecell{Voxel\\velocity} &
    \makecell{Ego-velocity\\(radar)} \\
    \midrule

    RadarOcc~\cite{radarocc}
    & CAS & None
    & free / background / foreground
    & -- & -- & -- \\

    \addlinespace[2pt]
    RaDelft detector~\cite{radelft}
    & CAS & None
    & binary (ground removed)
    & -- & -- & -- \\

    \addlinespace[2pt]
    Radar-Mamba~\cite{radarmamba}
    & CAS & None
    & binary (ground removed)
    & -- & -- & -- \\

    \addlinespace[2pt]
    AutoLabel~\cite{autolabel}
    & CAS & None
    & \makecell[l]{free / object classes\\(ground removed)}
    & \makecell[l]{scenario objects / pedestrian /\\
      vehicle / bicycle (no moving/static distinction)}
    & -- & -- \\

    \addlinespace[2pt]
    RIGS~\cite{rigs}
    & CAS & Camera + ego poses
    & free / background / foreground
    & -- & \checkmark & -- \\

    \midrule
    \multirow{2}{*}{\textbf{RadarOSV (ours)}}
    & CAS & Ego velocity
    & \multirow{2}{*}{free / ground / object}
    & \multirow{2}{*}{\makecell[l]{static / car / person /\\
      two-wheeler / large vehicle}}
    & \checkmark & -- \\

    \cmidrule(lr){2-3}\cmidrule(lr){6-7}
    & CAS + SC & None
    & & & \checkmark & \checkmark \\

    \bottomrule
  \end{tabular*}

  \par\vspace{2pt}
  \begin{minipage}{\textwidth}
    \scriptsize
    CAS: cascaded imaging radar; SC: single-chip radar. External inputs are those needed at inference.
    ``--'': no corresponding output; object classes exclude coarse occupancy labels.
    Dynamic-object classes follow annotations, not a speed threshold.
    The CAS-only row is a separately trained variant with the same outputs, not the occupancy-only PCG.
  \end{minipage}

\vspace{-2mm}
      
\end{table*}

\subsection{Camera- and LiDAR-Based Occupancy and Motion Perception}
\label{sec:rw_camera_lidar}

BEVFormer aggregates multi-camera features into a unified BEV space
for object detection and map perception~\cite{bevformer}.
3D occupancy prediction and semantic scene completion instead describe
geometry and semantics at the voxel level.
SemanticKITTI provides an outdoor benchmark,
while LMSCNet recovers multiscale geometry and semantics from voxelized
sparse LiDAR scans~\cite{semantickitti,lmscnet}.
MonoScene and SurroundOcc predict 3D occupancy from monocular and
multi-camera images, respectively, while Co-Occ improves reconstruction
through camera--LiDAR feature fusion and volume-rendering
constraints~\cite{monoscene,surroundocc,coocc}.

MotionNet predicts object categories, motion states, and future displacements in BEV from LiDAR sequences.
ViewFormer jointly predicts 3D occupancy and voxel-level motion from
temporal multi-view images, while Cam4DOcc forecasts future
occupancy~\cite{motionnet,viewformer,cam4docc}.
These methods learn motion mainly from temporal changes in visual or geometric data, whereas we use radar's direct radial-velocity measurements for voxel-level joint perception.

\subsection{Millimeter-Wave Radar Perception for Autonomous Driving}
\label{sec:rw_radar_perception}

Radar point-cloud methods include self-supervised scene flow estimation
with RaFlow, cross-modally supervised motion learning with CMFlow,
tracking with RaTrack, and joint moving-object segmentation and
ego-velocity estimation with
RadarMOSEVE~\cite{raflow,cmflow,ratrack,radarmoseve}.
4D-ROLLS learns occupancy representations from radar point clouds under
LiDAR supervision~\cite{4drolls}.
Doracamom fuses multi-view images and 4D radar point clouds for semantic
occupancy prediction and 3D detection with object-level velocity
estimation~\cite{doracamom}.
Radar point clouds thus support scene reconstruction, but sparsity and noise complicate modeling, and front-end detection such as CFAR may discard useful returns~\cite{radarocc}.

Another line of research directly processes radar spectra, heatmaps, or tensors for object detection and
classification, free-space and semantic segmentation~\cite{rodnet,raddet,fftradnet,tmvanet,cubedn}.
Radarize estimates translational and rotational ego-motion for robots, while RadarMP jointly predicts binary target/noise
segmentation and 3D scene flow from two consecutive radar
tensors~\cite{radarize,radarmp}.
RaMOD predicts dynamic-object categories
and velocities from short sequences of range--azimuth--Doppler (RAD)
tensors while estimating ego-motion, but its outputs are object-level~\cite{ramod}.

For multi-radar fusion, Pointillism exploits complementary spatial
observations from two same-model single-chip radars, using cross-radar
point-cloud consistency to suppress noise and estimate 3D boxes~\cite{pointillism}. Although ColoRadar provides cascaded and single-chip data,
SDDiff uses only its single-chip raw ADC measurements for point-cloud
extraction and ego-velocity estimation~\cite{coloradar,sddiff}.

\subsection{Radar-Tensor-Based Occupancy and Joint Perception}
\label{sec:rw_radar_occupancy}

Jin et al.\ learn 2D occupancy grids from range--Doppler data and
construct occupancy flow using grid predictions at consecutive time
steps to distinguish moving objects~\cite{jin_radar_occupancy}.
For 3D scenes, RadarOcc performs Doppler encoding and spatial feature
extraction on 4D imaging radar tensors, followed by spherical-to-Cartesian
feature aggregation to predict free, foreground, and background occupancy~\cite{radarocc}.

The RaDelft detector and Radar-Mamba take processed radar tensors and combine temporal information with Doppler features to
predict binary 3D occupancy under LiDAR-derived supervision.
Their predictions are then converted into enhanced radar
point clouds~\cite{radelft,radarmamba}.
Muckelroy et al.\ compare segmentation backbones and temporal modules for point-cloud quality within the RaDelft framework~\cite{muckelroy2025impact}.
Although these 3D methods use Doppler, they focus on geometric reconstruction rather than on dynamic-object semantic classes and voxel-level velocities.

AutoLabel and the concurrent RadarGSOcc learn radar voxel semantics from camera- or LiDAR-derived pseudo-labels, but do not distinguish moving from stationary objects or predict velocity; RadarGSOcc also finds road users hard to predict~\cite{autolabel,radargsocc}.
RIGS combines camera images, radar tensors, and ego poses within a
Gaussian scene representation to jointly predict occupancy, 3D voxel
velocities, and uncertainty.
Its displacement-guided Doppler disambiguation constructs velocity
pseudo-labels for training, while the model predicts a dense velocity
field at inference.
Inference still requires camera images and ego poses~\cite{rigs}.

In contrast, RadarOSV jointly predicts scene occupancy, dynamic-object
semantic classes, and ground-relative object velocity in unified
voxels. At inference, the full dual-radar configuration selects radial-velocity
candidates from SC Doppler spectra and estimates ego-velocity from radar
measurements, requiring no cameras, LiDAR, or external odometry.
Table~\ref{tab:capability} compares inference inputs and outputs.

\section{Method}
\label{sec:method}

RadarOSV combines complementary measurements from a cascaded imaging radar (CAS) and a single-chip radar (SC) for joint 3D scene and motion perception, predicting scene occupancy, dynamic object semantics, and their ground-relative velocity. After signal processing, CAS and SC produce a range-azimuth-elevation (RAE) cube $\mathbf X_t^{\mathrm{cas}}\in\mathbb R^{256\times256\times40}$ and a range-azimuth-Doppler (RAD) cube $\mathbf X_t^{\mathrm{sc}}\in\mathbb R^{192\times256\times255}$, respectively. Using the current frame and two preceding frames, the model produces geometry predictions $\hat{\mathbf y}$ for free, ground, and object classes on a $0.2\,\mathrm m$ Cartesian voxel grid, together with semantic scores $\hat{\mathbf s}_i$ and ground-relative velocity $\hat{\mathbf v}_i\in\mathbb R^2$ for object voxels.

\begin{figure*}[t]
    \centering
    \includegraphics[width=0.98\textwidth]{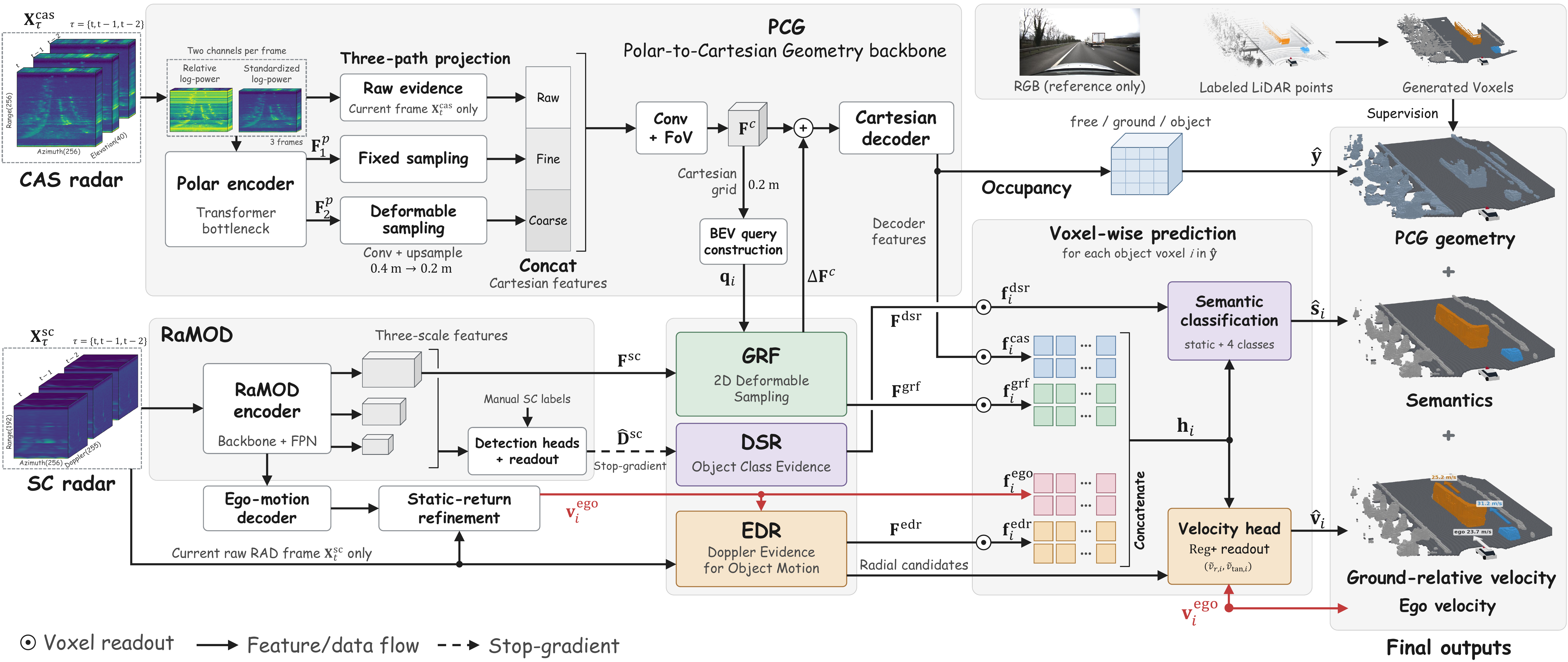}
    \caption{
        \textbf{RadarOSV architecture.}
        PCG constructs Cartesian geometry from CAS measurements,
        and GRF uses this geometry to aggregate SC features for
        occupancy and attribute prediction.
        DSR supplies class evidence only to the semantic head,
        while EDR provides motion features and radial-velocity candidates.
        Radar-estimated ego motion supports spectrum alignment
        and ground-relative velocity recovery.
        The dashed connection marks the stop-gradient boundary
        at the DSR input.
        RGB images are shown for reference only.
        LiDAR-derived voxels and manual SC labels provide
        training supervision.
    }
    \label{fig:radarosv_architecture}

    \vspace{-3mm}
    
\end{figure*}

As shown in Fig.~\ref{fig:radarosv_architecture}, the Polar-to-Cartesian Geometry (PCG) backbone first encodes CAS measurements and projects the resulting polar features into a 3D Cartesian feature volume $\mathbf F^c$. The Geometry-Guided Radar Fusion (GRF) module constructs bird's-eye view (BEV) queries from $\mathbf F^c$ to guide deformable aggregation of SC features. The aggregated features produce a geometric residual $\Delta\mathbf F^c$, which is added to $\mathbf F^c$ before the Cartesian decoder predicts 3D occupancy $\hat{\mathbf y}$. GRF also provides features $\mathbf F^{\mathrm{grf}}$ for subsequent semantic and velocity prediction.

The Explicit Doppler Readout (EDR) module extracts radial-velocity candidates and learned spectral features from SC Doppler spectra to support voxel-wise estimation of ground-relative object velocity. The Dynamic Semantic Readout (DSR) module provides semantic evidence from SC detection predictions. For each target voxel, we concatenate features from the Cartesian decoder, GRF, EDR, and ego-motion context into a fused feature vector $\mathbf h_i$, used by both the semantic classification and velocity regression heads. DSR features are additionally supplied to the classification head only. Throughout, $\mathbf F$ denotes an entire feature tensor, and $\mathbf f_i$ denotes the feature vector of target voxel $i$.

\subsection{Polar-to-Cartesian Geometry (PCG) Backbone}

To support joint prediction of geometry, semantics, and velocity in a common voxel space, PCG consists of a polar encoder and a Cartesian decoder linked by fixed and deformable sampling.

\textbf{Polar encoding.} For each input frame $\mathbf X_t^{\mathrm{cas}}$, we append a relative log-power channel to its standardized log-power to highlight contrast within each range bin. A 3D residual U-Net with a Transformer bottleneck jointly encodes three frames into fine- and coarse-scale features $\mathbf F_1^p$ and $\mathbf F_2^p$.

\textbf{Raw evidence and fixed sampling.} For each voxel center $\mathbf x_i$ in the target Cartesian grid, the fixed geometric mapping $\mathbf p_i=\Pi^{\mathrm{cas}}(\mathbf x_i)$ gives its corresponding sampling location in the CAS RAE grid. The raw-evidence path bypasses the encoder and samples the current-frame standardized log-power, while the fixed-sampling path samples $\mathbf F_1^p$. Both use trilinear interpolation at $\mathbf p_i$ to produce feature volumes on the $0.2\,\mathrm m$ Cartesian grid.

\textbf{Deformable sampling.} The coarse-scale features $\mathbf F_2^p$ are projected from RAE space onto a $0.4\,\mathrm m$ Cartesian grid through deformable sampling. For each target voxel center $\mathbf x_i$, we first sample $\mathbf F_2^p$ at the corresponding location $\mathbf p_i$ in the coarse RAE grid using trilinear interpolation. We concatenate the sampled feature with a positional encoding of the Cartesian coordinates $\mathbf x_i$ to form a query $\mathbf q_i$.

The query predicts $K$ local offsets in the RAE source grid and their aggregation weights. We then sample the original $\mathbf F_2^p$ at these shifted locations and combine the sampled features using the predicted weights. The deformable aggregation $\mathcal A$ produces a feature vector for each target location and takes a common form for PCG and GRF~\cite{deformable_detr}:

\begin{equation}
\mathcal A(\mathbf F,\mathbf q_i,\mathbf p_i)
=
\sum_{k=1}^{K}
a_k(\mathbf q_i)
\mathcal S\bigl(\mathbf F,\mathbf p_i+\boldsymbol\delta_k(\mathbf q_i)\bigr),
\tag{1}
\end{equation}

where $\mathbf F$ is the source feature tensor ($\mathbf F=\mathbf F_2^p$ in PCG), and $\mathcal S$ is trilinear sampling in PCG and bilinear sampling in GRF. $\boldsymbol\delta_k(\mathbf q_i)$ is the learned offset from $\mathbf p_i$ for the $k$-th sample, and $a_k(\mathbf q_i)$ is its normalized aggregation weight.

The Cartesian feature volume obtained from Eq.~(1) is upsampled from $0.4\,\mathrm m$ to $0.2\,\mathrm m$ and concatenated with the raw-evidence and fixed-sampling outputs along the channel dimension. These complementary features are fused by a convolution and masked to the radar field of view (FoV) to give $\mathbf F^c$, combining raw power evidence, fine-scale structure, and adaptively aggregated local context for Cartesian decoding and query construction in GRF.

\textbf{Cartesian decoding.} In the dual-radar model, the Cartesian decoder takes $\mathbf F^c+\Delta\mathbf F^c$ and produces geometry predictions $\hat{\mathbf y}$ at $0.2\,\mathrm m$ resolution. It assigns each voxel to free space, ground, or object and is supervised with LiDAR-derived voxel labels. The decoder also outputs a per-voxel feature vector $\mathbf f_i^{\mathrm{cas}}$ from its intermediate features, supporting subsequent fusion for semantic and velocity prediction.

\subsection{Cross-Radar Feature Fusion}

We initialize the SC branch with the pretrained RaMOD encoder and its detection and ego-motion readouts~\cite{ramod}. We extract features $\mathbf F^{\mathrm{sc}}$ from its fine-scale encoding layer and denote the decoded detection predictions by $\hat{\mathbf D}^{\mathrm{sc}}$. GRF uses the encoded features to provide cross-radar context, DSR adapts the class evidence in the detection predictions to voxel semantic prediction, and EDR extracts motion evidence from the raw SC spectra.

\textbf{Geometry-guided fusion (GRF).} We fold the Doppler dimension of $\mathbf F^{\mathrm{sc}}$ into the channel dimension to obtain source features on a range-azimuth grid. To select relevant SC features using CAS geometric context, we construct BEV queries $\mathbf q_i$ from the Cartesian feature volume $\mathbf F^c$ produced by PCG, before adding the cross-radar residual. We aggregate $\mathbf F^c$ along height and combine the resulting BEV features with positional encoding to form the queries. For each cell center in the target BEV grid, the fixed geometric mapping determines its corresponding sampling position $\mathbf p_i$ in the SC range-azimuth grid. The corresponding query $\mathbf q_i$ predicts local offsets and aggregation weights in this source grid. We then apply the two-dimensional form of Eq.~(1), using bilinear sampling at the shifted positions to aggregate SC features onto the Cartesian BEV grid.

GRF produces two outputs from these aggregated BEV features to support both geometry decoding and object-attribute prediction. One is a 3D residual $\Delta\mathbf F^c$, obtained by upsampling and expanding the features along height. This residual adds SC information to the CAS geometric representation, so that the Cartesian decoder receives $\mathbf F^c+\Delta\mathbf F^c$. The other is a BEV feature map $\mathbf F^{\mathrm{grf}}$. For each target voxel, we sample $\mathbf F^{\mathrm{grf}}$ at its BEV location to obtain $\mathbf f_i^{\mathrm{grf}}$, which is concatenated with $\mathbf f_i^{\mathrm{cas}}$ and the other branch features for semantic classification and velocity prediction.

\textbf{Dynamic Semantic Readout (DSR).} DSR converts SC detection predictions into spatial semantic features, allowing object-level class evidence to complement the shared voxel features. It extracts class confidence and object descriptors from the three-scale RaMOD predictions $\hat{\mathbf D}^{\mathrm{sc}}$ and fuses them on a common range-azimuth grid.

Using the same fixed geometric mapping as GRF, we project this representation onto the common Cartesian BEV grid to obtain $\mathbf F^{\mathrm{dsr}}$. For each target voxel, the feature $\mathbf f_i^{\mathrm{dsr}}$ is concatenated with the other voxel features only at the semantic classification head. Supplementary Section~S3 details the evidence aggregation and mapping process.

\textbf{Explicit Doppler Readout (EDR).} Doppler measurements constrain radial motion but do not directly provide the full two-dimensional velocity. EDR extracts this measurement evidence from the current SC RAD power cube before static suppression to complement learned motion features. We map the spectra onto the common Cartesian BEV grid and align them using the RaMOD ego-motion estimate, whose forward component is refined using static returns. This brings static responses close to zero velocity while preserving their energy.

EDR provides two complementary outputs. It combines learned spectral features with spectral descriptors to form $\mathbf F^{\mathrm{edr}}$. The corresponding voxel features $\mathbf f_i^{\mathrm{edr}}$ enter the subsequent fusion for semantic and velocity prediction. In parallel, EDR detects Doppler peaks, refines their locations, and converts them back to ego-relative radial-velocity candidates for the final velocity readout. Supplementary Section~S3 details ego-motion refinement, spectrum alignment, spectral encoding, and candidate extraction.

\subsection{Feature Fusion and Prediction Heads}

Semantic and velocity predictions are defined on the same $0.2\,\mathrm m$ Cartesian voxel grid as the geometry predictions $\hat{\mathbf y}$. For each voxel $i$, we concatenate the features $\mathbf f_i^{\mathrm{cas}}$, $\mathbf f_i^{\mathrm{grf}}$, and $\mathbf f_i^{\mathrm{edr}}$ from PCG, GRF, and EDR with an ego-motion context vector $\mathbf f_i^{\mathrm{ego}}$. This context contains the local radial and tangential components of ego velocity and the yaw rate, derived from the same ego-motion estimate used in EDR.

The fused vector $\mathbf h_i$ provides shared geometric and motion context for both prediction heads, while $\mathbf f_i^{\mathrm{dsr}}$ is additionally supplied only to the semantic classification head:
\begin{equation}
\begin{gathered}
\mathbf h_i=
[\mathbf f_i^{\mathrm{cas}};
 \mathbf f_i^{\mathrm{grf}};
 \mathbf f_i^{\mathrm{ego}};
 \mathbf f_i^{\mathrm{edr}}],\\
\hat{\mathbf s}_i=
\operatorname{Cls}([\mathbf h_i;\mathbf f_i^{\mathrm{dsr}}]),\\
(\tilde v_{r,i},\tilde v_{\mathrm{tan},i})
=\operatorname{Reg}(\mathbf h_i).
\end{gathered}
\tag{2}
\end{equation}

Here, $\operatorname{Cls}$ predicts scores for static, car, person, two-wheeler, and large vehicle. $\operatorname{Reg}$ predicts radial and tangential velocity components relative to the ego vehicle.

Conditioned on $\mathbf h_i$ and the regressed radial estimate, a learned selector scores $\tilde v_{r,i}$ alongside the EDR candidates expanded across neighboring Doppler ambiguity periods. At inference, it selects the highest-scoring candidate and defaults to $\tilde v_{r,i}$ when EDR measurements are invalid. The resulting radial component and $\tilde v_{\mathrm{tan},i}$ are rotated into Cartesian coordinates and combined with local ego-motion compensation to obtain the ground-relative velocity $\hat{\mathbf v}_i$. This combines radial measurements with learned tangential motion without restricting object motion to the forward direction. Supplementary Section~S3 details the velocity readout.

At inference, the two heads predict attributes for voxels classified as object by $\hat{\mathbf y}$. Velocities are set to zero for voxels assigned to the static semantic class.

\subsection{Supervision and Training}

LiDAR voxel and tracking labels supervise geometry, semantics, and ground-relative velocity. During training, object classes use LiDAR labels by default, with matched SC annotations taking priority. Valid SC radial-velocity annotations replace only the radial component, while the tangential component remains LiDAR-derived. SC annotations also supervise the RaMOD detection head. Evaluation follows the metric-specific annotation sources described in Sec.~\ref{sec:exp_setup}. At inference, the full dual-radar system uses its own radar-based ego-motion estimate.

The joint training objective is

\begin{equation}
\begin{gathered}
\mathcal L=\mathcal L_{\mathrm{geo}}+\lambda_{\mathrm{sem}}\mathcal L_{\mathrm{sem}}+\lambda_{\mathrm{vel}}\mathcal L_{\mathrm{vel}} \\
{}+\lambda_{\mathrm{sel}}\mathcal L_{\mathrm{sel}}+\lambda_{\mathrm{aux}}\mathcal L_{\mathrm{aux}}.
\end{gathered}
\tag{3}
\end{equation}

The geometry output of PCG and the final semantic classification head are supervised using weighted cross-entropy and Lovász~\cite{lovasz_softmax} losses. For voxels annotated as non-static, Smooth-L1 supervises the predicted radial and tangential velocity components, with additional supervision on the radial estimate produced by the velocity regression head. A candidate-selection loss trains the final velocity readout to select a suitable EDR radial candidate or use the regressed radial estimate, supervised by the ground-truth radial velocity. The auxiliary loss supervises the detection, velocity, and ego-motion predictions of the RaMOD branch.

We first train PCG for geometry prediction. The dual-radar model is then initialized with the trained PCG and pretrained RaMOD weights and jointly fine-tuned. We stop gradients at the RaMOD detection outputs used by DSR, while keeping the DSR adaptation layers trainable. During joint training, we randomly mask the SC contributions to the main prediction tasks for a subset of samples to reduce reliance on SC evidence. RadarOSV is designed for dual-radar input, while also supporting CAS-only and SC-only configurations with the same task outputs. These are trained separately, and their controlled input comparisons use external odometry to keep the ego-motion input consistent. Label-processing details, optimization settings, and single-radar configurations are provided in Supplementary Section~S2.

\section{Experiments}
\label{sec:experiments}

\subsection{Experimental Setup}
\label{sec:exp_setup}

\noindent\textbf{Datasets.} We evaluate PCG's geometric reconstruction on RaDelft~\cite{radelft} and K-Radar~\cite{kradar}, and the complete RadarOSV system on the synchronized dual-radar dataset BRIDGE-Liv, following the RaDelft, RadarOcc~\cite{radarocc}, and RaMOD~\cite{ramod} splits, respectively. On BRIDGE-Liv, occupancy, class, and velocity labels come from LiDAR point clouds and object tracks, while manual SC annotations are the primary reference for object radial velocity.

\noindent\textbf{Metrics.} Geometry mIoU averages the ground and object IoUs, and dynamic-object IoU is computed over voxels annotated as dynamic objects, regardless of class. Object class and velocity are evaluated within annotated object regions. Velocity errors are mean absolute errors (MAE), and planar speed is the magnitude of the ground-relative planar velocity. See Supplementary Sections~S1--S2.

\noindent\textbf{Configurations.} Unless otherwise stated, both radars use three input frames, with two power channels per CAS frame. The PCG evaluated on BRIDGE-Liv is trained jointly on BRIDGE-Liv and RaDelft, whereas public-dataset comparisons use separately trained models. The complete system takes its ego velocity from the RaMOD estimate, refined in the forward direction using static returns.

\subsection{PCG Geometry Results on Public Datasets}
\label{sec:exp_public}

We first evaluate PCG alone against published results.

\noindent\textbf{RaDelft.} With its default output, PCG reaches a 3D Chamfer distance of 1.176\,m, lower than both the RaDelft detector (1.537\,m) and Radar-Mamba~\cite{radarmamba} (1.228\,m) (Table~\ref{tab:radelft}). With the object-probability threshold at 0.30, which closely matches the detector's false-alarm probability, PCG achieves both a higher detection probability (67.5\% vs.\ 62.1\%) and a lower Chamfer distance (1.269 vs.\ 1.537\,m).

\begin{table}[!t]
  \centering
  \caption{Geometry results on RaDelft. $P_d$ and $P_{fa}$ denote detection and false-alarm probabilities, respectively. NR: Not Reported.}
  \label{tab:radelft}
  \footnotesize
  \setlength{\tabcolsep}{3pt}
  \renewcommand{\arraystretch}{1.08}
  \begin{tabular*}{\columnwidth}{@{\extracolsep{\fill}}lccc@{}}
    \toprule
    Method / setting & \makecell{3D Chamfer\,$\downarrow$\\(m)} & $P_d\uparrow$ & $P_{fa}\downarrow$ \\
    \midrule
    RaDelft detector~\cite{radelft} & 1.537 & 62.1\% & 2.77\% \\
    Radar-Mamba~\cite{radarmamba} & 1.228 & NR & NR \\
    PCG (default) & 1.176 & 50.2\% & 1.24\% \\
    PCG (threshold 0.30) & 1.269 & 67.5\% & 2.73\% \\
    \bottomrule
  \end{tabular*}
\end{table}

\noindent\textbf{K-Radar.} To match RadarOcc~\cite{radarocc}, PCG uses a single input frame with two power channels and follows its 0.4\,m grid and evaluation ranges. PCG outperforms RadarOcc on all four metrics at all three ranges (Table~\ref{tab:kradar}), on a test split that includes night, rain, and heavy-snow sequences. These results support PCG as the spatial representation for class and motion prediction.

\begin{table}[!t]
  \centering
  \caption{Occupancy results on K-Radar (\%, higher is better) with single-frame input. In each cell, the first number is RadarOcc and the second, in bold, is PCG. BG/FG: background/foreground.}
  \label{tab:kradar}
  \footnotesize
  \setlength{\tabcolsep}{2.5pt}
  \renewcommand{\arraystretch}{1.08}
  \begin{tabular*}{\columnwidth}{@{\extracolsep{\fill}}lcccc@{}}
    \toprule
    & \multicolumn{4}{c}{RadarOcc~\cite{radarocc} / \textbf{PCG (ours)}} \\
    \cmidrule(l){2-5}
    Range & IoU & mIoU & BG IoU & FG IoU \\
    \midrule
    12.8\,m & 48.8 / \textbf{52.9} & 34.3 / \textbf{40.7} & 47.9 / \textbf{52.7} & 20.7 / \textbf{28.8} \\
    25.6\,m & 39.1 / \textbf{44.7} & 28.5 / \textbf{34.2} & 38.2 / \textbf{44.3} & 18.7 / \textbf{24.2} \\
    51.2\,m & 30.4 / \textbf{33.8} & 22.6 / \textbf{27.7} & 29.4 / \textbf{32.8} & 15.8 / \textbf{22.6} \\
    \bottomrule
  \end{tabular*}
  
\end{table}

\subsection{Complete RadarOSV System}
\label{sec:exp_system}

The last row of Table~\ref{tab:system} reports the complete system on BRIDGE-Liv: a single model that jointly outputs geometry, object classes, object velocity, and ego velocity from the two radars (examples in Fig.~\ref{fig:radarosv_teaser}). The first row is its PCG initialization; rows N1--N5 are analyzed in Sec.~\ref{sec:exp_ablation}.

\begin{table*}[!t]
  \centering
  \caption{Results on BRIDGE-Liv (IoU and accuracy in \%, MAE in m/s). PCG is the first-stage geometry network. N1--N5 are separately trained configurations. RadarOSV is the complete system. CAS-only adds class and velocity heads to PCG; SC-only masks the CAS input.}    
  \label{tab:system}
  \footnotesize
  \setlength{\tabcolsep}{3pt}
  \renewcommand{\arraystretch}{1.12}
  \begin{tabular*}{\textwidth}{@{\extracolsep{\fill}}lccc*{6}{c}@{}}
    \toprule
    Configuration & \makecell{CAS\\frames} & \makecell{SC\\frames} & \makecell{Ego\\velocity} & \makecell{Geometry\\mIoU\,$\uparrow$} & \makecell{Object\\IoU\,$\uparrow$} & \makecell{Dynamic-object\\IoU\,$\uparrow$} & \makecell{Object-class\\macro accuracy\,$\uparrow$} & \makecell{Object radial\\MAE\,$\downarrow$\\Manual (m/s)} & \makecell{Object planar-speed\\MAE\,$\downarrow$\\LiDAR (m/s)} \\
    \midrule
    PCG & 3 & -- & -- & 29.8 & 16.0 & -- & -- & -- & -- \\
    \midrule
    N1 CAS-only & 1 & -- & Odometry & 29.8 & 17.6 & 23.2 & 56.3 & 2.484 & 2.244 \\
    N2 CAS-only & 3 & -- & Odometry & 30.7 & 18.9 & 26.2 & 63.1 & 1.025 & 0.988 \\
    N3 SC-only & -- & 3 & Odometry & 18.2 & 3.9 & 16.6 & 75.8 & 0.443 & 0.837 \\
    N4 CAS+SC & 1 & 3 & Odometry & 29.8 & 17.7 & 23.3 & 75.9 & 0.450 & 0.797 \\
    N5 CAS+SC & 3 & 3 & Odometry & 30.8 & 18.8 & 26.1 & 74.6 & 0.428 & 0.737 \\
    \midrule
    \textbf{RadarOSV} & 3 & 3 & Radar & 30.7 & 18.8 & 25.6 & 75.5 & 0.505 & 0.801 \\
    \bottomrule
  \end{tabular*}
  \par\smallskip
  \begin{minipage}{\textwidth}
    \footnotesize
    N3--N5 and RadarOSV additionally use manual SC labels for matched objects during training (Supplementary Section~S2); Table~\ref{tab:readout} isolates the SC modules under identical supervision. Over five training runs, RadarOSV obtains a class macro accuracy of $74.7\pm0.6$\% and an object radial-velocity MAE of $0.500\pm0.008$\,m/s (mean $\pm$ standard deviation); the selection of the reported run is described in Supplementary Section~S5.
  \end{minipage}
\end{table*}

\noindent\textbf{Geometry.}
RadarOSV preserves the reconstruction quality of its PCG initialization, with slightly higher mIoU (30.7\% vs.\ 29.8\%) and object IoU (18.8\% vs.\ 16.0\%); range-binned results are in Supplementary Section~S8.

\noindent\textbf{Object class.}
RadarOSV reaches 75.5\% macro accuracy over the four object classes (car 95.5\%, person 64.4\%, two-wheeler 60.8\%, large vehicle 81.5\%). The contributions of the individual SC modules are analyzed in Sec.~\ref{sec:exp_ablation}.

\noindent\textbf{Object and ego velocity.}
At the same annotated object locations and against the same manual Doppler annotations, RadarOSV reaches an object radial-velocity MAE of 0.505\,m/s, less than half of the 1.098\,m/s of the original RaMOD model re-evaluated on them (Supplementary Section~S4). The object planar-speed MAE against LiDAR tracks, which also involves the tangential prediction and ego-motion compensation, is 0.801\,m/s. Across separately trained configurations, static-return refinement reduces ego-speed MAE from 0.430 to 0.160\,m/s against LiDAR--IMU odometry and object radial-velocity MAE from 0.622 to 0.505\,m/s (Supplementary Sections~S4). The complete system requires no external odometry at inference.

\noindent\textbf{Efficiency.}
On a single RTX 5090 (FP16, batch size 1), the complete system, including PCG and the SC branch, runs at 12.5 FPS (80.2\,ms per frame), 30.5\,ms more than PCG alone (49.7\,ms); on an RTX 3090 it takes 174.0\,ms (5.7 FPS). Details are given in Supplementary Section~S5.

\begin{figure*}[t]
  \centering
  \includegraphics[width=\textwidth]{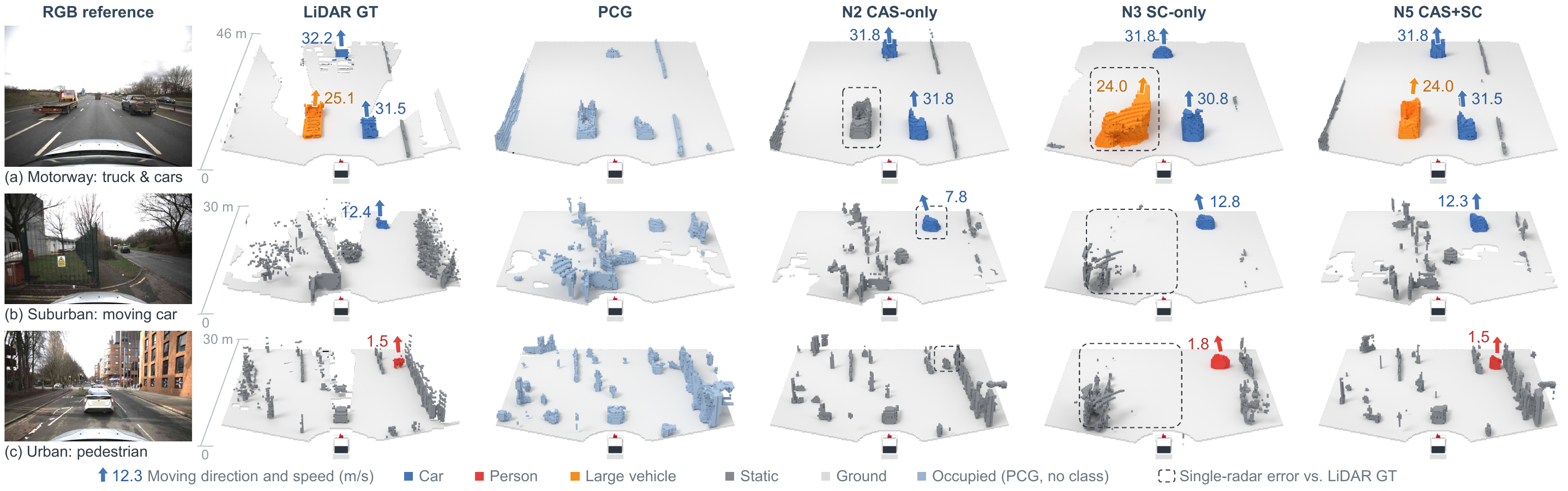}
  \caption{\textbf{Input ablation on three BRIDGE-Liv test scenes.}
  Columns: RGB reference, LiDAR ground truth, first-stage PCG geometry, and the separately trained CAS-only (N2), SC-only (N3), and CAS+SC (N5) configurations of Table~\ref{tab:system} (external odometry).
  Dashed boxes mark single-radar errors with respect to the LiDAR ground truth that N5 resolves.
  (a)~N2 gives the truck no consistent class and N3 spreads it over about three times its footprint; N5 recovers class and extent.
  (b)~N2 underestimates the car's speed (7.8 vs.\ 12.4\,m/s) and N3 misses the fence; N5 gives 12.3\,m/s and keeps the fence.
  (c)~N2 shows the pedestrian as static and N3 misses the structures ahead, including the stopped vehicle; N5 recovers both.
  Speeds as in Fig.~\ref{fig:radarosv_teaser}; each object component takes its dominant class, or static if none dominates.}
  \label{fig:input_ablation}
  
\end{figure*}

\subsection{Ablation Studies}
\label{sec:exp_ablation}

\noindent\textbf{PCG input frames and power channels.} Under the same joint training, using three CAS frames instead of one improves geometry mIoU from 29.1\% to 29.8\% on BRIDGE-Liv and from 25.8\% to 26.5\% on RaDelft, and reduces the RaDelft Chamfer distance from 1.49 to 1.37\,m, confirming the benefit of temporal context. Adding the relative log-power channel, which enhances echo contrast within each range bin, raises the five-run mean class accuracy from 72.3\% to 74.7\% (Supplementary Sections~S5 and S6).

\noindent\textbf{Complementarity of the two radars.} Rows N1--N5 of Table~\ref{tab:system} compare separately trained CAS-only, SC-only, and CAS+SC configurations under the same external odometry (initialization in Supplementary Section~S2). Compared with N2, the dual-radar configuration N5 raises class accuracy from 63.1\% to 74.6\% and reduces radial-velocity MAE from 1.025 to 0.428\,m/s and planar-speed MAE from 0.988 to 0.737\,m/s, while maintaining geometry performance. Conversely, SC alone yields only 3.9\% object IoU (N3), versus 18.8\% for N5. Among the evaluated configurations, only CAS+SC combines strong geometric reconstruction with accurate object classification and velocity estimation. Fig.~\ref{fig:input_ablation} shows the same pattern qualitatively: N2 has errors in class and velocity, N3 on object shape and static structure, and N5 resolves both. Without SC input, using three CAS frames instead of one (N1 vs.\ N2) reduces radial-velocity MAE from 2.484 to 1.025\,m/s, showing that consecutive geometric observations also carry motion cues.

\noindent\textbf{GRF.} Table~\ref{tab:readout} adds the SC modules one at a time to the CAS-only model N2; all rows are trained separately with the same PCG initialization, LiDAR main-task supervision, and external odometry. Fusing encoded SC features into the CAS representation through GRF raises class macro accuracy from 63.1\% to 68.4\% and reduces radial-velocity MAE from 1.025 to 0.745\,m/s while maintaining geometry performance, supporting cross-radar feature fusion.

\begin{table}[!t]
  \centering
  \caption{Contribution of GRF, EDR, and DSR. Each row adds one module to the row above, starting from CAS-only N2. Metrics and units follow Table~\ref{tab:system}: radial MAE against manual Doppler annotations and planar-speed MAE against LiDAR tracks. All rows use LiDAR main-task supervision and external odometry.}
  \label{tab:readout}
  \footnotesize
  \setlength{\tabcolsep}{2.5pt}
  \renewcommand{\arraystretch}{1.08}
  \begin{tabular*}{\columnwidth}{@{\extracolsep{\fill}}lccccc@{}}
    \toprule
    Configuration & \makecell{Geom.\\mIoU\,$\uparrow$} & \makecell{Dyn.\\IoU\,$\uparrow$} & \makecell{Class\\macro acc.\,$\uparrow$} & \makecell{Radial\\MAE\,$\downarrow$} & \makecell{Planar\\MAE\,$\downarrow$} \\
    \midrule
    CAS-only (N2) & 30.7 & 26.2 & 63.1 & 1.025 & 0.988 \\
    $+$SC features (GRF) & 30.9 & 26.2 & 68.4 & 0.745 & 0.839 \\
    $+$EDR & 30.7 & 25.7 & 70.1 & 0.521 & 0.715 \\
    $+$DSR & 30.7 & 25.7 & 73.9 & 0.519 & 0.712 \\
    \bottomrule
  \end{tabular*}
\end{table}

\noindent\textbf{EDR.} Adding EDR further reduces radial-velocity MAE from 0.745 to 0.521\,m/s and planar-speed MAE from 0.839 to 0.715\,m/s (Table~\ref{tab:readout}). Spectral features and radial candidates read from the raw spectrum thus provide motion evidence beyond the encoded SC features. The $+$EDR row adds both EDR outputs; with the complete model's weights fixed, replacing the final candidate selection by the continuous regression alone raises radial-velocity MAE from 0.505 to 0.819\,m/s (Supplementary Section~S4).

\noindent\textbf{DSR.} Adding DSR on top of EDR raises class macro accuracy from 70.1\% to 73.9\%, with the largest gain for persons (58.6\% to 65.6\%; Supplementary Section~S4). Geometry and velocity errors remain largely unchanged, consistent with DSR supplying class evidence to the semantic head.

\section{Conclusion}
\label{sec:conclusion}

We presented RadarOSV, to our knowledge the first radar-only framework that jointly estimates 3D occupancy, object classes, and ground-relative velocity from raw radar data in unified voxels, with radar-based ego-motion estimation replacing external odometry at inference. Its PCG backbone outperforms the compared radar methods on RaDelft and K-Radar. On BRIDGE-Liv, geometry-guided fusion with explicit Doppler and dynamic semantic readouts raises object-class accuracy from 63.1\% to 73.9\% and nearly halves the radial-velocity error relative to the same network without them, at unchanged geometry performance. Future work will extend it to single-radar multi-mode and Doppler-division multiplexing radars, reduce latency, and integrate it into end-to-end planning.

\section*{LLM Usage Statement}

The research idea, the problem formulation, and the method were proposed by the authors, who made all design decisions and conducted the research. LLM-based assistants (Claude, Anthropic; Codex, OpenAI) were used as engineering and editing aids. They assisted with implementing and debugging the code, running experiments, and literature search. All code and outputs produced with their help were reviewed and verified by the authors. The authors wrote the manuscript, with the assistants helping to revise and polish the drafts in all sections. All reported results and figures come from real experiments on real data; none was produced by an LLM or other generative AI tool. The authors take full responsibility for all content. Code and model checkpoints will be released upon acceptance.

{
    \small
    \bibliographystyle{ieeenat_fullname}
    \bibliography{ref}
}

\clearpage
\newcounter{pagebeforesupp}
\setcounter{pagebeforesupp}{\value{page}}
\let\origmaketitlesupplementary\maketitlesupplementary
\renewcommand{\maketitlesupplementary}{\setcounter{page}{\value{pagebeforesupp}}\origmaketitlesupplementary}

\let\origfigure\figure
\let\endorigfigure\endfigure
\newcommand{\suppfigsfivesix}{%
\begin{origfigure}[!t]
  \centering
  \includegraphics[width=\columnwidth]{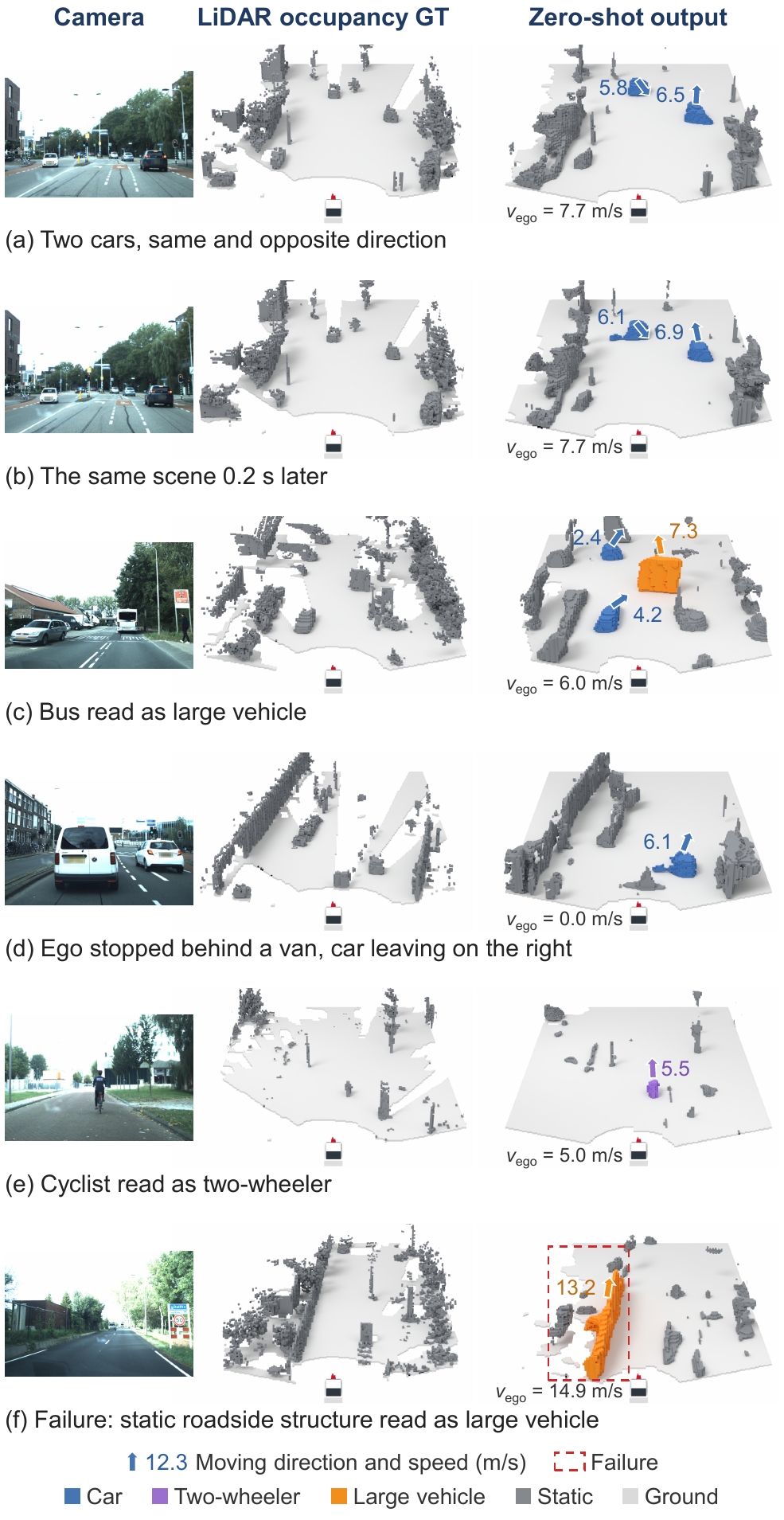}
  \caption{Zero-shot transfer of class and velocity to RaDelft test scenes with N2 (CAS only).
  Each row shows the camera image for reference, the LiDAR occupancy ground truth, and the model output.
  The ground truth has no class labels, so all its objects are shown in gray.
  Arrows and numbers show the predicted velocity over the ground.
  (a, b) Two cars, one moving in the same direction as the ego vehicle and one in the opposite direction, are predicted as car in two frames 0.2\,s apart, with consistent class and velocity.
  (c) A bus is predicted as large vehicle.
  (d) The ego vehicle and the van ahead are stopped. The van stays static, and the car leaving on the right is predicted as car.
  (e) A cyclist is predicted as two-wheeler.
  (f) Failure case. A static roadside structure is predicted as large vehicle.
  Each connected component is shown in a single class.}
  \label{fig:supp_zeroshot}
\end{origfigure}
\begin{origfigure}[!t]
  \centering
  \includegraphics[width=\columnwidth]{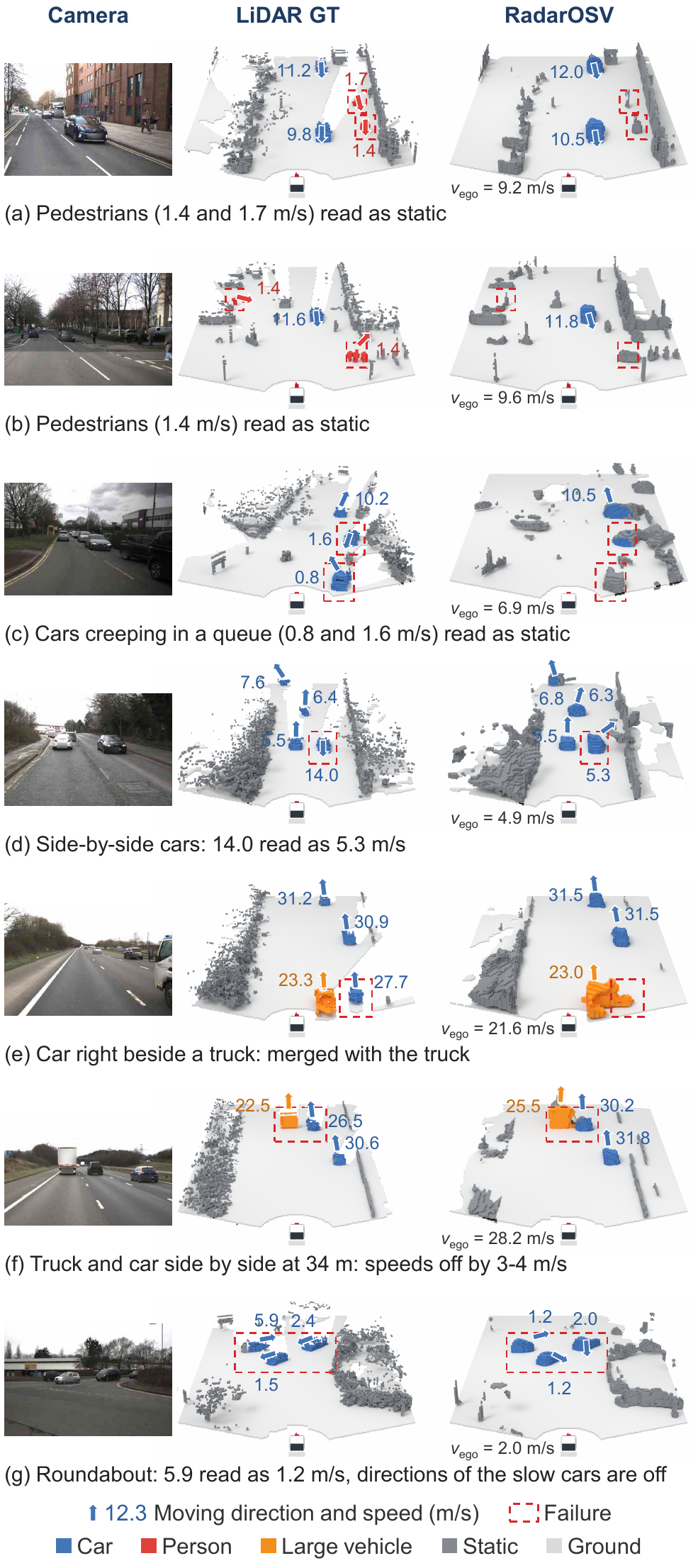}
  \caption{Failure cases on BRIDGE-Liv.
  The camera images are for reference, and red dashed boxes mark the failed objects.
  (a--c) Slow objects. (d--f) Neighboring objects. (g) Tangential motion.
  Arrows and numbers show the velocity of each object, taken as the median over its voxels (Sec.~\ref{sec:supp_eval}). An object predicted as static has no arrow.
  For display, a connected component is drawn in a single moving class when that class makes up most of it, a moving class that makes up a very small part is not drawn, and all other voxels are drawn as predicted.}
  \label{fig:supp_failure}
\end{origfigure}}
\newcounter{suppcolfig}
\RenewDocumentEnvironment{figure}{O{t} +b}{%
  \stepcounter{suppcolfig}%
  \ifnumcomp{\value{suppcolfig}}{<}{2}{\begin{origfigure}[#1]#2\end{origfigure}}{}%
}{}
\let\origsection\section
\newcounter{suppsec}
\renewcommand{\section}{\stepcounter{suppsec}%
  \ifnum\value{suppsec}=7 \afterpage{\clearpage\suppfigsfivesix\clearpage}\fi\origsection}

\clearpage
\setcounter{page}{1}
\maketitlesupplementary

\setcounter{section}{0}
\setcounter{figure}{0}
\setcounter{table}{0}
\setcounter{equation}{0}
\renewcommand{\thesection}{S\arabic{section}}
\renewcommand{\thefigure}{S\arabic{figure}}
\renewcommand{\thetable}{S\arabic{table}}
\renewcommand{\theequation}{S\arabic{equation}}
\renewcommand{\theHsection}{S\arabic{section}}
\renewcommand{\theHfigure}{S\arabic{figure}}
\renewcommand{\theHtable}{S\arabic{table}}
\renewcommand{\theHequation}{S\arabic{equation}}
\section{Radar Configurations and Datasets}
\label{sec:supp_radar_data}

\subsection{Datasets and Radar Configurations}
\label{sec:supp_datasets}

RadarOSV uses four datasets that provide raw radar data. Table~\ref{tab:supp_datasets} lists the details of these datasets, including the radar models, the annotations, and the model input tensor sizes. LiDAR and cameras are used for label generation, evaluation, and visualization, whereas RadarOSV uses only radar data at inference.

\begin{table*}[!t]
  \centering
  \caption{Radar configurations, model inputs, and object annotations of the four datasets.}
  \label{tab:supp_datasets}
  \footnotesize
  \setlength{\tabcolsep}{2pt}
  \renewcommand{\arraystretch}{1.12}
  \begin{tabular*}{\textwidth}{@{\extracolsep{\fill}}llccclll@{}}
    \toprule
    Dataset & Radar & \makecell{$v_{\max}$\\(m/s)} & Range & \makecell{Dual-radar\\synchronization} & Model input & Object annotations \\
    \midrule
    K-Radar~\cite{kradar} & RETINA-4ST & $\approx\pm1.9$ & 118\,m & -- & RAE, $256\times107\times37$ & \makecell[l]{3D boxes annotated on LiDAR and\\transformed to radar (class)} \\
    RaDelft~\cite{radelft} & TI MMWCAS & $\pm2.49$ & $\approx$51\,m & -- & RAE, $256\times256\times40$ & \makecell[l]{Automatic semantic labels on LiDAR~\cite{autolabel}\\(class, moving and static not distinguished)} \\
    ColoRadar~\cite{coloradar} & \makecell[l]{TI MMWCAS /\\AWR1843} & \makecell{$\pm2.02$ /\\$\pm2.56$} & \makecell{15\,m /\\12\,m} & \makecell{Software\\timestamps} & RAE, $256\times256\times40$ & None \\
    BRIDGE-Liv~\cite{ramod} & \makecell[l]{TI MMWCAS /\\AWR2243} & \makecell{$\pm3.38$ /\\$\pm23.2$} & \makecell{49\,m /\\49\,m} & \makecell{Hardware\\trigger} & \makecell[l]{RAE, $256\times256\times40$ /\\RAD, $192\times256\times255$} & \makecell[l]{Manual LiDAR object tracks (class, ground-relative\\velocity) and manual SC boxes (class, radial velocity)} \\
    \bottomrule
  \end{tabular*}
  \par\smallskip
  \begin{minipage}{\textwidth}
    \footnotesize
    For datasets with two radars, entries are given as cascaded / single-chip. The single-chip radar of ColoRadar is not used. Only BRIDGE-Liv offers two hardware-synchronized radars, a $\pm23.2$\,m/s velocity range, and manual annotations of object class and velocity.
  \end{minipage}
\end{table*}

PCG relies only on the imaging radar, so it can be trained and evaluated on all four datasets. The complete RadarOSV also requires a complementary radar that provides additional Doppler and classification cues. Its training and evaluation further require class and velocity annotations. Among the four datasets, only BRIDGE-Liv provides both, so we evaluate class and velocity on BRIDGE-Liv and compare only the geometry on RaDelft and K-Radar (Tables~2 and~3 of the main paper).

Although K-Radar provides 3D boxes that are annotated on LiDAR point clouds and transformed to the radar frame, these boxes have class labels but no velocity or Doppler annotations. RaDelft provides no manual annotations. Its semantic labels are point-wise LiDAR labels generated automatically from camera and LiDAR data by AutoLabel~\cite{autolabel}, and they do not separate moving and static objects or provide velocities. On 50 manually labeled frames, the pedestrian labels reach a precision of only 0.67 and a recall of 0.72~\cite{autolabel}. RadarOSV assigns classes only to moving objects, so these labels cannot be used to evaluate its class and velocity outputs. Table~\ref{tab:supp_frames} lists the number of training, validation, and test frames.

\begin{table}[!t]
  \centering
  \caption{Number of training, validation, and test frames.}
  \label{tab:supp_frames}
  \footnotesize
  \setlength{\tabcolsep}{3pt}
  \renewcommand{\arraystretch}{1.08}
  \begin{tabular*}{\columnwidth}{@{\extracolsep{\fill}}lccc@{}}
    \toprule
    Dataset / stage & Training & Validation & Test \\
    \midrule
    BRIDGE-Liv, PCG training & 58,509 & 6,529 & 10,367 \\
    BRIDGE-Liv, joint training & 37,342 & 3,571 & 10,367 \\
    RaDelft & 12,175 & 1,352 & 5,020 \\
    K-Radar & 11,360 & 1,059 & 2,878 \\
    ColoRadar & 10,645 & -- & 2,404 \\
    \bottomrule
  \end{tabular*}
  \par\smallskip
  \begin{minipage}{\columnwidth}
    \footnotesize
    The test sets follow existing splits, namely the RaMOD split~\cite{ramod} for BRIDGE-Liv, the official test scenes for RaDelft, and the RadarOcc split~\cite{radarocc} for K-Radar. Validation frames of BRIDGE-Liv and RaDelft are held out from the training data. Joint training on BRIDGE-Liv uses only frames with manual radar annotations. ColoRadar is used only for the transfer experiments in Sec.~S6 and has no validation set.
  \end{minipage}
\end{table}

\subsection{Roles of the Two Radars}
\label{sec:supp_roles}

RadarOSV uses two millimeter-wave radars to combine the high angular resolution of a cascaded imaging radar (CAS, 12 transmit and 16 receive antennas) with the wide velocity range of a single-chip radar (SC, 2 transmit and 4 receive antennas). These two properties are linked by a basic trade-off in time-division-multiplexed (TDM) MIMO radar.

An FMCW radar measures radial velocity from the phase change between consecutive chirps. In a TDM-MIMO radar, the transmit antennas transmit in turn, so consecutive chirps from the same antenna are $N_{\mathrm{TX}}$ chirp periods $T_c$ apart. The maximum unambiguous radial velocity is therefore
\begin{equation}
v_{\max}=\frac{\lambda}{4\,N_{\mathrm{TX}}\,T_c},
\label{eq:supp_vmax}
\end{equation}
where $\lambda$ is the wavelength.

\begin{figure*}[t]
  \centering
  \includegraphics[width=\textwidth]{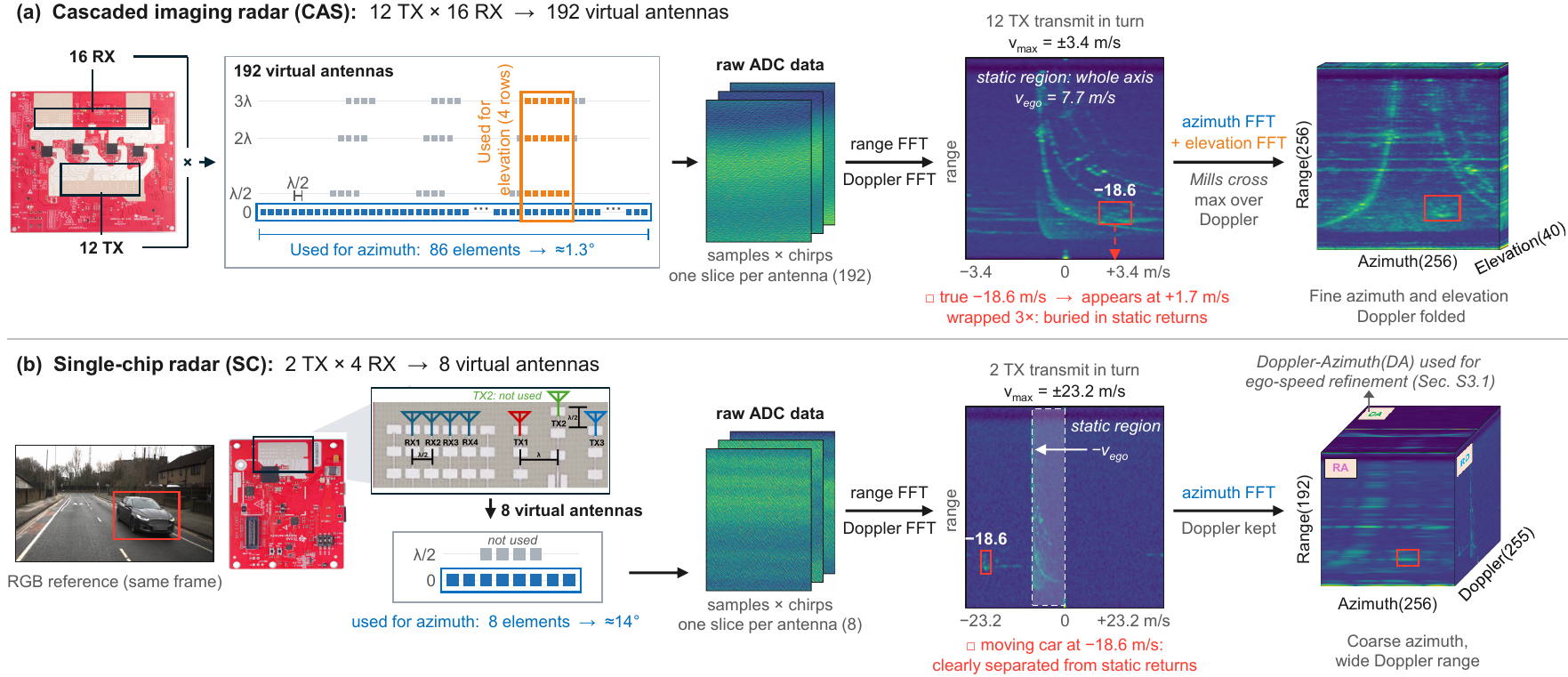}
  \caption{From antennas to input cubes.
  Both radars observe the same frame, with the ego vehicle moving at 7.7\,m/s.
  The red boxes mark an oncoming car.
  The RGB image is for reference only.}
  \label{fig:supp_radars}
\end{figure*}

More transmit antennas give a larger virtual aperture and a finer angular resolution, but reduce $v_{\max}$ in proportion to $N_{\mathrm{TX}}$. Fig.~\ref{fig:supp_radars} shows both radars on the same frame. CAS forms 192 virtual antennas, of which 86 contiguous elements are used for azimuth and four rows for elevation. With $N_{\mathrm{TX}}=12$, $T_c=24\,\mu\mathrm{s}$, and $\lambda\approx3.9$\,mm, however, its $v_{\max}$ is only about $\pm3.38$\,m/s. SC forms eight virtual antennas, all along azimuth. It cannot resolve elevation, but with $N_{\mathrm{TX}}=2$, its $v_{\max}$ is about $\pm23.2$\,m/s.

The two radars therefore have opposite strengths. In the range-azimuth-elevation (RAE) cube of CAS, the oncoming car appears as a compact object in azimuth, and the walls on both sides are clearly visible. In the range-azimuth-Doppler (RAD) cube of SC, the same car is spread along azimuth, while in Doppler, SC is the clearer radar. When the ego vehicle moves, static objects also have a radial velocity, which falls between $-v_{\mathrm{ego}}$ and 0 in the SC range-Doppler map. The car has a relative radial velocity of $-18.6$\,m/s and lies outside this interval, so its velocity can be read directly. On CAS, the car wraps three times and appears at $+1.7$\,m/s, and the static returns also wrap repeatedly and cover the whole Doppler axis. Recovering the object velocity from CAS would require separating the object from the static returns in the same Doppler bin using azimuth, and then finding the true velocity among several ambiguity periods. The same holds for K-Radar and RaDelft, whose $v_{\max}$ is only about $\pm1.9$ and $\pm2.5$\,m/s (Table~\ref{tab:supp_datasets}). RadarOSV therefore uses CAS for spatial structure and takes all Doppler velocity measurements from SC.

RaMOD~\cite{ramod} reports that different classes of road users have distinct Doppler patterns, which are important cues for detection and classification. This is consistent with Table~5 of the main paper, where adding the three SC modules one at a time to the CAS-only model raises the object-class macro accuracy from 63.1\% to 73.9\%.

\subsection{Form of the Input Cubes}
\label{sec:supp_input}

For both radars, range and Doppler FFTs of the raw ADC data produce a range-Doppler map, and an angle FFT then produces the input cube (Fig.~\ref{fig:supp_radars}). For CAS, azimuth and elevation are combined as in a Mills cross~\cite{mills_cross}, and the Doppler dimension is reduced by max pooling, which keeps the return energy at each location. Table~\ref{tab:supp_datasets} lists the cube sizes. Both radars use three consecutive frames as input, and EDR additionally reads the raw spectrum of the current SC frame. Each CAS frame has two power channels. The first is the standardized log-power. The second is a relative log-power obtained by subtracting the maximum log-power within the same range bin, which highlights the contrast among returns at the same range.

Keeping range, azimuth, elevation, and Doppler together would give $256\times256\times40\times128\approx3.4\times10^{8}$ cells per CAS frame. This is about 1.3\,GB in FP32 and about 4\,GB for three frames, before any intermediate network features are counted. For comparison, the 4D tensors of K-Radar take about 260\,MB per frame and about 12\,TB for the whole dataset. Existing methods therefore compress the input first. The RaDelft detector~\cite{radelft} keeps only the strongest elevation bin for each range-azimuth-Doppler cell, so its input is a $500\times240\times128$ RAD power tensor with the corresponding elevation indices. RadarOcc~\cite{radarocc} summarizes the 64 Doppler bins at each location into eight statistics and keeps only the strongest elements at each range.

RadarOSV instead takes two 3D cubes. The RAE cube from CAS keeps the full spatial resolution (about 2.62 million cells), and the RAD cube from SC keeps the full Doppler spectrum (about 12.53 million cells). Together, they are about 1/22 the size of the 4D CAS tensor. The two cubes share the range-azimuth plane, where they can be aligned and fused.

\subsection{Independence of the Architecture from Dual-Radar Hardware}
\label{sec:supp_independence}

RadarOSV requires two complementary observations, an imaging cube with high spatial resolution and a Doppler cube with a sufficiently wide velocity range. Using two physical radars is only one way to obtain them.

\noindent\textbf{Two subframes of one cascaded radar.} TI cascaded radars can set the transmit antennas for each chirp and divide a frame into several subframes~\cite{ti_awr2243_cascade}. A single radar can therefore acquire an imaging subframe with 12 transmit and 16 receive antennas for the RAE cube, followed by a velocity subframe with 1 transmit and 16 receive antennas for the RAD cube. By Eq.~(\ref{eq:supp_vmax}), the $v_{\max}$ of the velocity subframe is 12 times larger. The two subframes take about 43\,ms in total, which fits within a 10\,FPS frame period, and the network needs no change.

\noindent\textbf{DDM radar.} In a Doppler-division-multiplexing (DDM) radar, all transmit antennas operate at the same time, and all virtual channels are obtained from the same chirps. This gives a 4D tensor over range, azimuth, elevation, and Doppler. The tensor is again too large (Sec.~\ref{sec:supp_input}), so the same approach applies. It is split into an RAE cube and an RAD cube, which are fed to the two branches.

We are currently studying RadarOSV with a single multi-mode radar and with DDM radars.

\section{Label Generation, Evaluation, and Training}
\label{sec:supp_labels}

\subsection{Geometry Labels from LiDAR}
\label{sec:supp_geo_labels}

The geometry labels of BRIDGE-Liv and RaDelft are generated from single LiDAR frames on a 0.2\,m Cartesian voxel grid that covers 0 to 51.2\,m forward, $\pm51.2$\,m to the left and right, and $-2.6$ to 3.0\,m in height ($256\times512\times28$ voxels). For BRIDGE-Liv, the point clouds are de-skewed with FAST-LIO2~\cite{fastlio2}, the points of moving objects are shifted to the end of the scan using their track velocity, and the ground is segmented with Patchwork++~\cite{patchworkpp}. For RaDelft, we use the point clouds provided by the dataset and take the ground segmentation from its ground-removed version.

\begin{figure*}[t]
  \centering
  \includegraphics[width=\textwidth]{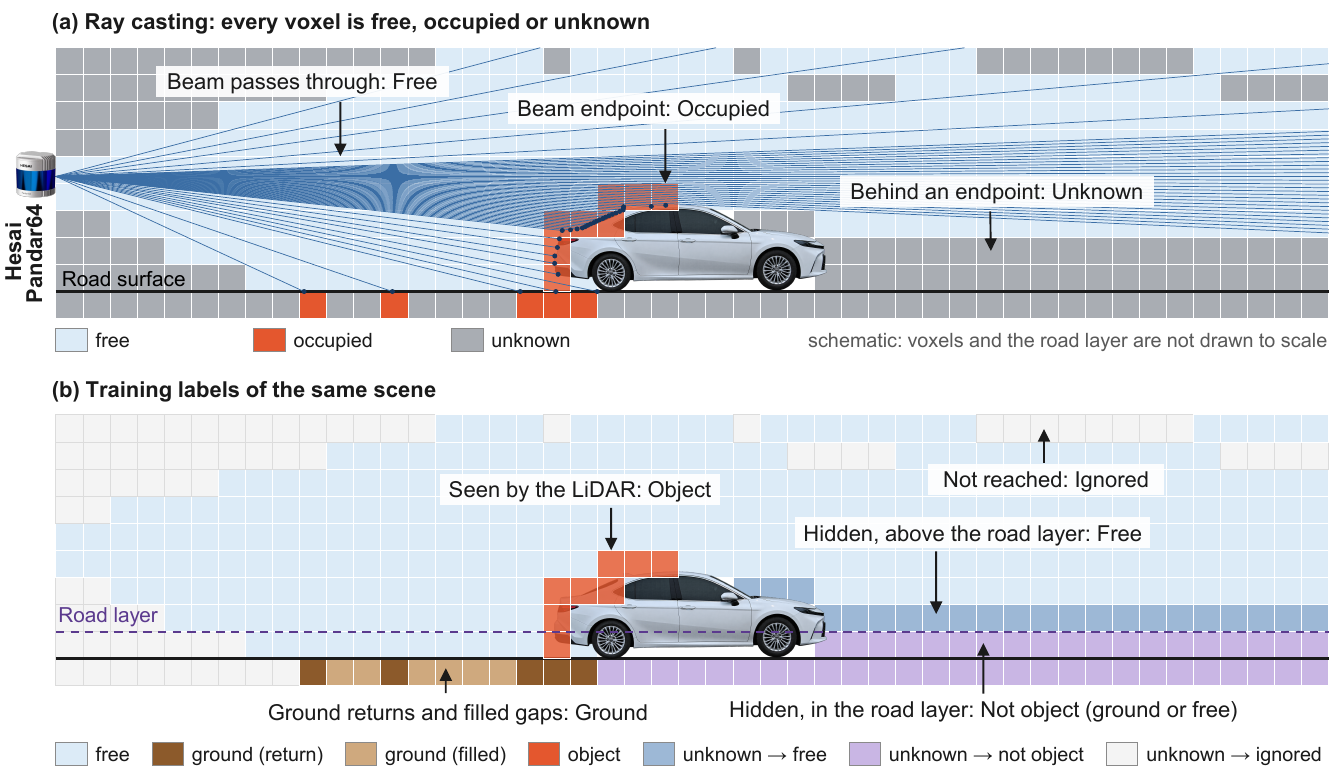}
  \caption{LiDAR ray casting and voxel labels (side view).
  (a)~The three voxel states after ray casting.
  (b)~Training labels of the same scene.
  The car is directly ahead and shown in a vertical section.
  For a car at another azimuth, the beams also hit its side that faces the LiDAR, and those voxels are labeled as object as well.}
  \label{fig:supp_labels}
\end{figure*}

The labels are generated in two steps (Fig.~\ref{fig:supp_labels}). The first step is ray casting. Along each beam, the voxels between the sensor and the return are free, the voxel that contains the return is occupied, and voxels that no beam passes through are unknown.

The second step converts these states into training labels. Occupied voxels are labeled as ground or object according to the ground segmentation, and the gaps between neighboring ground returns are filled as ground. The network predicts only free, ground, and object, while unknown voxels are either excluded from training or used as weak supervision, depending on the cause.

Some voxels are unknown because no beam reaches them. They lie either in the gaps between beams or outside the LiDAR coverage. Whether they contain an object cannot be determined, so they are excluded from training (light gray in Fig.~\ref{fig:supp_labels}b).

The other unknown voxels are occluded by an object or by the road surface. Leaving them unconstrained would let the network predict many false objects where the LiDAR cannot see. We treat them according to their height, using a road layer defined as the 0.8\,m above the ground. Occluded voxels above the road layer are supervised as free (darker blue). Occluded voxels inside the road layer may be either road or free space, because the exact road height cannot be determined from LiDAR. They are only constrained not to be object, and the network decides between ground and free (purple).

These rules affect only voxels that the LiDAR does not see. Object surfaces are supervised as object even inside the road layer. Unknown voxels are not scored in evaluation.

\subsection{Object Class and Velocity Labels}
\label{sec:supp_obj_labels}

Object class and ground-relative planar velocity labels come from manually annotated LiDAR objects and their tracks. Only moving objects have a class label. Stationary vehicles belong to the static class, such as those marked in Fig.~1(c) of the main paper.

BRIDGE-Liv also provides boxes manually annotated on the SC radar data~\cite{ramod}, each with a class and a radial velocity. We refer to them as SC annotations and use them in two ways during training. First, they supervise the detection head of the RaMOD branch. Second, they update the voxel labels of the main tasks. When a LiDAR object is matched to an SC annotation, its voxels take the SC class and, if valid, the SC radial velocity, while the tangential velocity still comes from the LiDAR track. Unmatched objects keep their LiDAR labels. All rows of Table~5 and rows N1 and N2 of Table~4 in the main paper use LiDAR labels for the main tasks, whereas N3--N5 and the complete system use the SC-updated labels.

\subsection{Evaluation Protocol}
\label{sec:supp_eval}

On RaDelft, the 3D Chamfer distance, $P_d$, and $P_{fa}$ follow the RaDelft protocol~\cite{radelft}. The ground truth is the ground-removed LiDAR point cloud provided by the dataset. For the Chamfer distance, the prediction is represented by the centers of the voxels predicted as object. $P_d$ and $P_{fa}$ are computed on the official evaluation grid, whose range cells (0.1\,m) are finer than our voxels (0.2\,m), so each predicted voxel is represented by eight points inside it. On K-Radar, we generate 0.4\,m labels with the RadarOcc scripts~\cite{radarocc}, use the same evaluation ranges, and select the checkpoint by validation mIoU.

On BRIDGE-Liv, object class and velocity are evaluated per object on the 13,306 moving objects in the test set that have an SC annotation, the same objects as in the RaMOD evaluation~\cite{ramod}. Objects that occupy fewer than five voxels are excluded. For each object, we take the most frequent predicted class and the median predicted velocity over its voxels and compare them with the annotations. Class accuracy is macro-averaged over the four classes.

\subsection{Training Settings}
\label{sec:supp_training}

RadarOSV is trained in two stages. The first stage trains only PCG for geometry and can use all training frames of BRIDGE-Liv and RaDelft. Class and velocity prediction also needs SC data and object annotations, so the second stage loads the trained PCG and the pretrained RaMOD and trains all tasks jointly on the annotated BRIDGE-Liv frames (Table~\ref{tab:supp_frames}).

Both stages use AdamW~\cite{adamw} with a weight decay of 0.01, a OneCycle learning-rate schedule~\cite{onecycle} with warm-up over the first 10\% of the iterations, 5 epochs, 4 RTX 5090 GPUs, and FP16 mixed precision, which is also used for evaluation. PCG is trained from scratch with a learning rate of $2\times10^{-4}$ and 3 samples per GPU, and its checkpoint is selected by validation mIoU. The joint stage uses a learning rate of $1\times10^{-4}$ ($5\times10^{-5}$ for the pretrained RaMOD branch) and 2 samples per GPU, and we take the weights of the last epoch.

The loss follows Eq.~(3) of the main paper with $\lambda_{\mathrm{sem}}=1$, $\lambda_{\mathrm{vel}}=0.1$, $\lambda_{\mathrm{sel}}=0.2$, and $\lambda_{\mathrm{aux}}=1$. The geometry loss has three terms,
\begin{equation*}
\mathcal L_{\mathrm{geo}}
=\mathcal L_{\mathrm{CE}}+\mathcal L_{\mathrm{Lov}}
+\frac{1}{|\mathcal U|}\sum_{i\in\mathcal U}-\log\bigl(1-p_i^{\mathrm{obj}}\bigr).
\end{equation*}
The first two terms are the weighted cross-entropy and Lov\'asz losses~\cite{lovasz_softmax} on labeled voxels, with class weights of 1, 2, and 3 for free, ground, and object. $\mathcal U$ is the set of occluded voxels inside the road layer (Sec.~\ref{sec:supp_geo_labels}), which are constrained only by the third term to have a low object probability $p_i^{\mathrm{obj}}$. $\mathcal L_{\mathrm{sem}}$ has the same form as the first two terms, with class weights of 1, 3, 6, 6, and 3 for static, car, person, two-wheeler, and large vehicle. The class weights balance the number of voxels per class. The target of $\mathcal L_{\mathrm{sel}}$ is the candidate closest to the ground-truth radial velocity, or the regressed value if no candidate lies within 0.75\,m/s. Gradients are stopped where DSR reads the detection outputs and where EDR picks the peaks, and the main-task losses are not back-propagated to the ego-motion estimate.

\noindent\textbf{Random SC masking.} During joint training, the SC contribution to the main tasks is switched off for each sample with a probability of 0.25. All features that GRF, EDR, and DSR pass to the decoder and the two heads are set to zero, so the decoder input is the same as in PCG alone. The losses of the RaMOD branch are still computed. This prevents the geometry and attribute predictions from relying too heavily on SC.

\noindent\textbf{Single-radar configurations.} CAS-only (N1, N2) removes the SC branch together with GRF, EDR, and DSR, and adds the class and velocity heads to PCG. SC-only (N3) keeps the full network, sets the CAS input cube to zero, and does not load the PCG weights, so all scene information comes from SC through the GRF residual. N1 and N4 are initialized from the single-frame PCG, and N2 and N5 from the three-frame PCG. These configurations are trained separately with LiDAR--IMU odometry as the ego-motion input, which removes the effect of ego-motion estimation errors. The complete system uses the ego motion estimated and refined by the radar itself.

\section{Additional Method Details}
\label{sec:supp_readout}

The main paper gives the full formulation of GRF. This section describes the refinement of the ego forward speed, the internal structure of EDR and DSR, and the details of the velocity and semantic classification heads. The corresponding ablations are in Sec.~S4.

\subsection{Refinement of the Ego Forward Speed}
\label{sec:supp_ego_refine}

The ego-motion decoder of the RaMOD branch estimates the translational velocity and the yaw rate of the ego vehicle, $(v_x, v_y, \omega_z)$, from the SC data. Unlike RaMOD, RadarOSV further refines the forward speed $v_x$ using the current SC frame, because an error in $v_x$ affects the ground-relative velocity of every object.

\begin{figure}[t]
  \centering
  \includegraphics[width=\columnwidth]{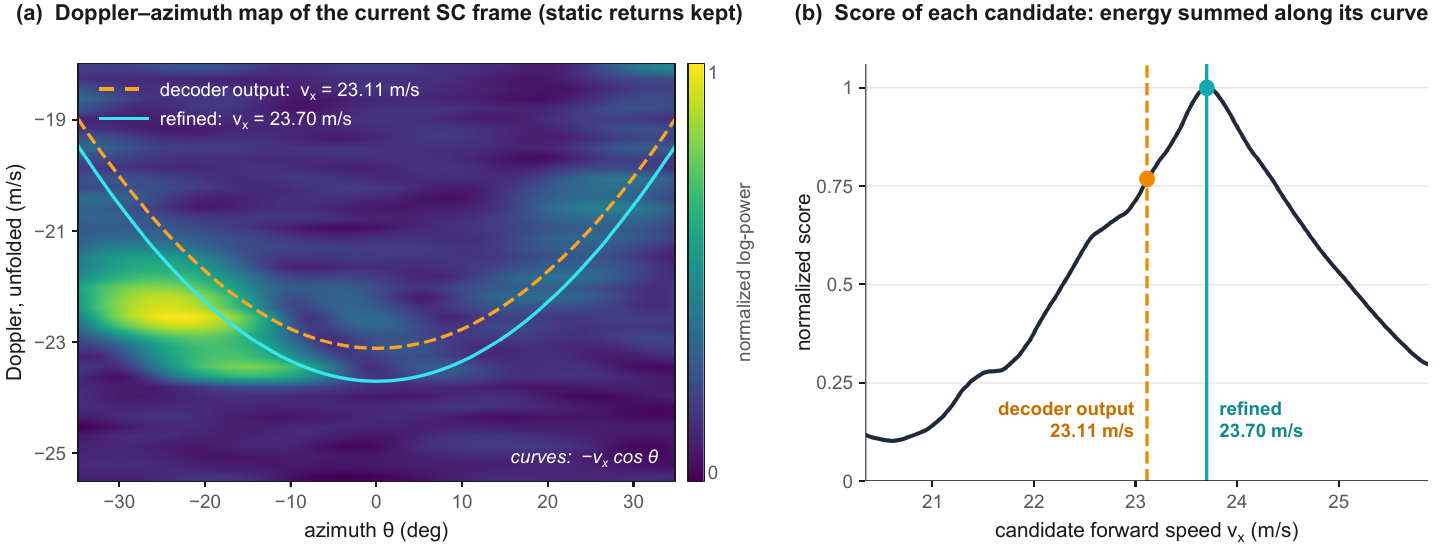}
  \caption{Refinement of the ego forward speed on real data.}
  \label{fig:supp_ego_refine}
\end{figure}

A static object at azimuth $\theta$ has a Doppler of $-v_x\cos\theta$~\cite{ramod}, so the static returns lie on a cosine curve in the Doppler-azimuth (DA) map. We then test a set of candidate speeds around the decoder output (within $\pm12\%$ and at least $\pm1$\,m/s). Each candidate defines a cosine curve, and its score is the energy summed along this curve. The candidate with the highest score is the refined $v_x$ (Fig.~\ref{fig:supp_ego_refine}). In the example, the decoder output is 23.11\,m/s and the refined value is 23.70\,m/s, while LiDAR--IMU odometry gives 23.64\,m/s. $v_y$ and $\omega_z$ are taken directly from the decoder.

The refinement uses only returns beyond 15\,m and within $\pm35^\circ$ in azimuth, because the road surface at closer range lies below the radar and would bias the estimate low. The refined speed is used for the spectrum alignment in EDR, the per-voxel ego-motion context $\mathbf f_i^{\mathrm{ego}}$, and the conversion to ground-relative velocity.

\subsection{Explicit Doppler Readout (EDR)}
\label{sec:supp_edr}

\begin{figure*}[t]
  \centering
  \includegraphics[width=\textwidth]{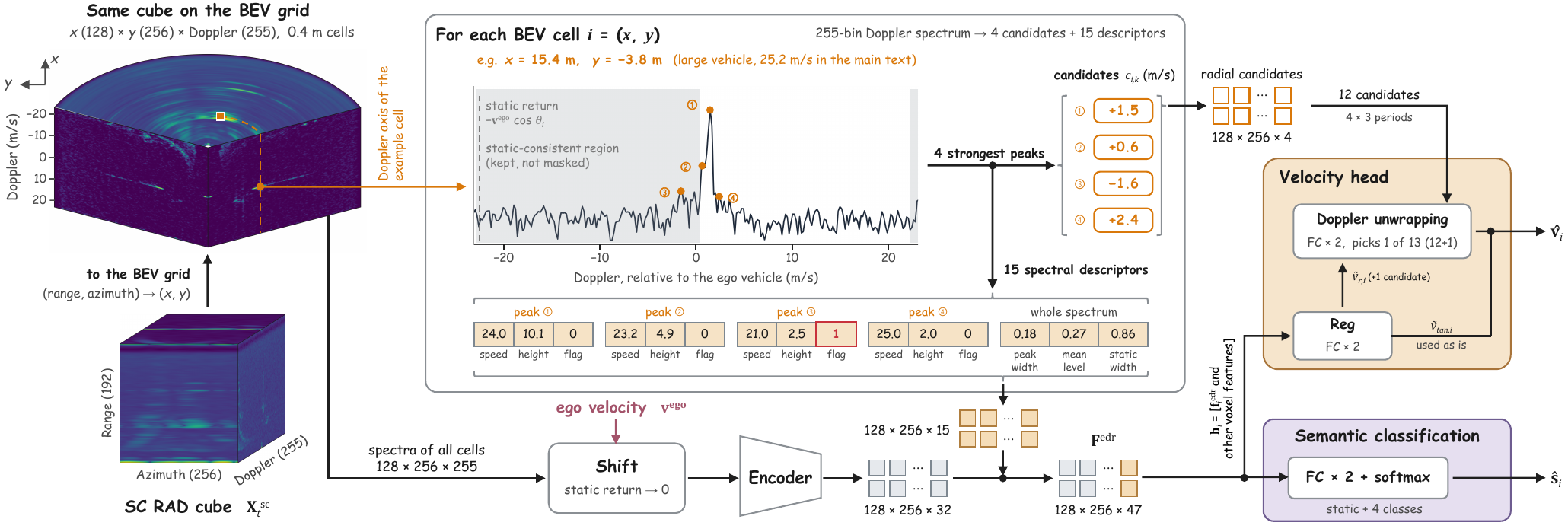}
  \caption{Details of the EDR readout on real data (sequence 3, frame 5402, the same frame as Fig.~2 of the main paper).
  Left, the SC RAD cube is resampled to the BEV grid. The orange square is the example cell, the dashed line is its Doppler axis (drawn on the side face at the range of this cell), and the dot is peak~1.
  Middle, the Doppler spectrum of the example cell with its four strongest peaks and the static-consistent Doppler region in gray.
  Bottom, gray squares are the encoder output and orange squares are the spectral descriptors.
  Right, the two prediction heads (Secs.~\ref{sec:supp_vel_readout} and~\ref{sec:supp_dsr}).}
  \label{fig:supp_edr}
\end{figure*}

The SC features fused by GRF have been downsampled in Doppler and passed through static suppression. They provide useful context but no longer carry fine velocity readings. RaMOD~\cite{ramod} suppresses the static-consistent Doppler region and then reads object velocity from features with a velocity head. EDR instead reads the velocity directly from the raw spectrum (Fig.~\ref{fig:supp_edr}).

The raw RAD cube of the current SC frame is first resampled from range-azimuth coordinates to the BEV grid, which gives a $128\times256\times255$ tensor with the Doppler dimension unchanged. For each BEV cell, EDR takes the four strongest peaks along the Doppler axis, at least four bins apart, and refines their positions by parabolic interpolation. The peaks are converted to radial velocities relative to the ego vehicle ($+1.5$, $+0.6$, $-1.6$, and $+2.4$\,m/s in the example cell) and passed to the velocity head as radial candidates. Some peaks fall inside the static-consistent Doppler region (gray in the figure, containing peak~3). Unlike RaMOD, we do not use this region and simply take the four strongest peaks.

EDR also records three quantities for each peak, namely its velocity relative to the static return (speed), its standardized energy (height), and a flag that marks whether it falls inside the static-consistent Doppler region (1 for peak~3, shown in red). Three more quantities describe the whole spectrum, namely the width of the main peak, the mean energy, and the spread of the static returns caused by ego motion. Together they form 15 spectral descriptors, similar to the Doppler bins descriptor of RadarOcc~\cite{radarocc}.

Object velocity is also a cue for object class, so EDR provides a feature path as well. The spectra of all cells are shifted along the Doppler axis using the refined ego speed (Sec.~\ref{sec:supp_ego_refine}) so that the static returns fall at zero, and are then encoded by three convolution layers. The encoder output is concatenated with the 15 spectral descriptors to form $\mathbf F^{\mathrm{edr}}$, which is passed to both prediction heads.

\subsection{Velocity Readout and Doppler Unwrapping}
\label{sec:supp_vel_readout}

The $v_{\max}$ of SC is $\pm23.2$\,m/s, so the relative radial velocity of an oncoming vehicle can exceed this range and wrap on the spectrum. The velocity head first regresses the radial and tangential velocities from the fused feature $\mathbf h_i$ with two fully connected layers (Reg in Fig.~\ref{fig:supp_edr}). The regressed value uses multi-frame geometry and motion information from CAS and is not limited to $\pm v_{\max}$, so it does not wrap, but its precision is limited. The four EDR candidates are precise but may be wrapped. We consider one wrap in each direction, so adding and subtracting one Doppler period (46.4\,m/s) to each candidate gives 12 candidates. Another two fully connected layers score these 12 candidates together with the regressed radial velocity, and the one with the highest score is the final radial velocity. In short, the regressed value decides the ambiguity period, and the spectral peak provides the precision. When the SC measurement is invalid, only the regressed value is used.

Doppler measures only the radial component, so the tangential component is learned from spatio-temporal features. The selected radial component $v_{r,i}$ and the regressed tangential component $\tilde v_{\mathrm{tan},i}$ are rotated to the output coordinates and added to the ego motion at voxel $i$, which gives the ground-relative velocity
\begin{equation}
\begin{gathered}
\hat{\mathbf v}_i=\mathbf R(\theta_i)\,[\,v_{r,i},\ \tilde v_{\mathrm{tan},i}\,]^{\top}+\mathbf v_i^{\mathrm{ego}},\\
\mathbf v_i^{\mathrm{ego}}=[\,v_x-\omega_z y_i,\ v_y+\omega_z x_i\,]^{\top},
\end{gathered}
\label{eq:supp_ground_vel}
\end{equation}
where $\theta_i$ and $(x_i, y_i)$ are the azimuth and the position of voxel $i$. This formulation does not restrict object motion to the forward direction, so it also applies to turning vehicles.

\subsection{Dynamic Semantic Readout (DSR)}
\label{sec:supp_dsr}

LiDAR voxel labels are attached to the surfaces that the LiDAR sees, whereas the RaMOD detector learns to classify objects from the radar echo structure itself. DSR uses the detector output as additional evidence for voxel classification.

DSR reads the predictions $\hat{\mathbf D}^{\mathrm{sc}}$ of the RaMOD detection head at three scales before NMS. At each range-azimuth-Doppler position, the detection head outputs a box (center and size along range, azimuth, and Doppler), an objectness confidence, and four class probabilities. DSR takes the maximum over Doppler and keeps nine quantities for each range-azimuth cell. These are the highest score of each of the four classes (confidence times class probability), the highest confidence, and the three sizes and the Doppler position of the highest-scoring box. The three scales are upsampled to the finest resolution, concatenated, and converted to the BEV grid by an encoder, which gives $\mathbf F^{\mathrm{dsr}}$.

$\mathbf F^{\mathrm{dsr}}$ is concatenated with $\mathbf h_i$ only at the input of the semantic classification head, which consists of two fully connected layers and a softmax. It is not passed to the velocity head. Gradients are stopped at the readout, so the voxel classification loss does not change the detection head, which remains supervised by the SC annotations.

\section{Additional Ablation Studies}
\label{sec:supp_readout_exp}

This section reports the ablations for the modules in Sec.~\ref{sec:supp_readout}. Results labeled complete system use the ego motion estimated and refined by the radar. N5 and the configurations in Table~\ref{tab:supp_per_class} use LiDAR--IMU odometry to keep the ego-motion input consistent for a controlled comparison of the modules.

\subsection{Effect of the Ego-Speed Refinement}
\label{sec:supp_exp_ego}

Table~\ref{tab:supp_ego_refine} compares the complete system with and without the refinement of Sec.~\ref{sec:supp_ego_refine}. The refinement reduces the ego-speed MAE from 0.430 to 0.160\,m/s, which is 37\% of the original error, and the object radial-velocity MAE from 0.622 to 0.505\,m/s. A more accurate ego speed thus gives more accurate object velocities. The refinement adds no learnable parameters and uses only the static returns of the current frame.

\begin{table}[!t]
  \centering
  \caption{Effect of the ego forward-speed refinement on two separately trained complete systems (MAE, m/s).}
  \label{tab:supp_ego_refine}
  \footnotesize
  \setlength{\tabcolsep}{3pt}
  \renewcommand{\arraystretch}{1.08}
  \begin{tabular*}{\columnwidth}{@{\extracolsep{\fill}}lcc@{}}
    \toprule
    Ego motion & Ego speed & \makecell{Object radial\\velocity} \\
    \midrule
    RaMOD ego estimate & 0.430 & 0.622 \\
    $+$ static-return refinement & \textbf{0.160} & \textbf{0.505} \\
    \bottomrule
  \end{tabular*}
\end{table}

\subsection{Selection of the Final Radial Velocity}
\label{sec:supp_exp_radial}

Table~\ref{tab:supp_radial_readout} examines the velocity head of Sec.~\ref{sec:supp_vel_readout}. With the model weights fixed, we change only how the final radial velocity is obtained. The first row gives the original RaMOD model for reference. It is re-evaluated on the same objects with its original protocol, which reads its velocity output at each manually annotated box, whether or not the object is detected.

\begin{table}[!t]
  \centering
  \caption{Object radial-velocity MAE (m/s) for different ways of obtaining the final radial velocity, with fixed weights.}
  \label{tab:supp_radial_readout}
  \footnotesize
  \setlength{\tabcolsep}{3pt}
  \renewcommand{\arraystretch}{1.08}
  \begin{tabular*}{\columnwidth}{@{\extracolsep{\fill}}lcc@{}}
    \toprule
    Method / final radial velocity & \makecell{Radar ego\\motion} & \makecell{External\\odometry} \\
    \midrule
    RaMOD~\cite{ramod} & 1.098 & 1.019 \\
    \midrule
    \textbf{Reg $+$ Doppler unwrapping (1 of 13)} & \textbf{0.505} & \textbf{0.428} \\
    Reg only & 0.819 & 0.679 \\
    4 candidates $+$ Reg, no period shift & 0.549 & 0.478 \\
    12 candidates, without Reg & 0.534 & 0.479 \\
    Strongest peak & 1.302 & 1.236 \\
    \bottomrule
  \end{tabular*}
  \par\smallskip
  \begin{minipage}{\columnwidth}
    \footnotesize
    For RadarOSV, the two columns are the complete system and N5. Evaluated on the 13,291 objects with a valid manual Doppler annotation.
  \end{minipage}
\end{table}

The complete velocity head gives the lowest error under both ego-motion sources, less than half of the RaMOD error, and using only the regressed value is clearly worse. Removing the candidates shifted by one Doppler period increases the error, which shows that these candidates resolve the wrapping. Removing the regressed value from the candidates also increases the error, which shows that it is needed as a fallback. Taking the strongest peak directly gives the largest error, so peak strength alone cannot determine the object velocity.

\subsection{Effect of EDR and DSR on Object Class}
\label{sec:supp_exp_class}

Table~\ref{tab:supp_per_class} adds EDR and DSR in turn to N2 $+$ GRF and reports the accuracy of each class.

\begin{table}[!t]
  \centering
  \caption{Per-class accuracy (\%) when EDR and DSR are added in turn.}
  \label{tab:supp_per_class}
  \footnotesize
  \setlength{\tabcolsep}{3pt}
  \renewcommand{\arraystretch}{1.08}
  \begin{tabular*}{\columnwidth}{@{\extracolsep{\fill}}lccccc@{}}
    \toprule
    Configuration & Car & Person & \makecell{Two-\\wheeler} & \makecell{Large\\vehicle} & \makecell{Macro\\average} \\
    \midrule
    N2 $+$ GRF & 95.1 & 53.2 & 51.9 & 73.4 & 68.4 \\
    $+$EDR & 95.6 & 58.6 & 53.2 & 73.2 & 70.1 \\
    $+$DSR (with EDR) & 94.8 & \textbf{65.6} & \textbf{59.5} & \textbf{75.6} & \textbf{73.9} \\
    \bottomrule
  \end{tabular*}
\end{table}

EDR raises the macro average from 68.4\% to 70.1\%, with the largest gain for persons (53.2\% to 58.6\%). Its spectral features describe how an object moves, and persons move most differently from vehicles. DSR further raises the macro average to 73.9\% and improves persons (58.6\% to 65.6\%), two-wheelers (53.2\% to 59.5\%), and large vehicles (73.2\% to 75.6\%) together. Voxel-wise classification sees only local features, whereas DSR adds the class decision of the detection head for the whole object. This helps most for small objects with few voxels and for large vehicles that are only partly visible. Averaged over car, person, and large vehicle only, the accuracy also rises from 75.8\% to 78.7\%.

\section{Repeated-Training Statistics and Runtime}
\label{sec:supp_runs}

Table~\ref{tab:supp_five_runs} reports the statistics over five training runs. RadarOSV reaches a class macro accuracy of $74.7\pm0.6$\% and an object radial-velocity MAE of $0.500\pm0.008$\,m/s, which shows that the results are stable. Adding the relative log-power channel to the CAS input raises the mean class macro accuracy from 72.3\% to 74.7\% and reduces its standard deviation from 2.2 to 0.6 percentage points.

\begin{table*}[!t]
  \centering
  \caption{Statistics over five training runs.}
  \label{tab:supp_five_runs}
  \footnotesize
  \setlength{\tabcolsep}{3pt}
  \renewcommand{\arraystretch}{1.12}
  \begin{tabular*}{\textwidth}{@{\extracolsep{\fill}}lcccc@{}}
    \toprule
    CAS input & \makecell{Class macro accuracy (\%)\\mean $\pm$ std} & \makecell{Class macro accuracy (\%)\\min--max} & \makecell{Object radial MAE (m/s)\\mean $\pm$ std} & \makecell{Object radial MAE (m/s)\\min--max} \\
    \midrule
    Standardized log-power & $72.3\pm2.2$ & 69.5--74.3 & $0.502\pm0.009$ & 0.493--0.515 \\
    $+$ relative log-power within each range bin & $\mathbf{74.7\pm0.6}$ & \textbf{73.9--75.5} & $\mathbf{0.500\pm0.008}$ & \textbf{0.491--0.509} \\
    \bottomrule
  \end{tabular*}
  \par\smallskip
  \begin{minipage}{\textwidth}
    \footnotesize
    Complete system with radar ego-motion estimation and refinement. Each row uses its own fixed PCG initialization, and each run is evaluated at the last epoch. The RadarOSV result in the main paper is the run with the highest class accuracy in the second row (75.5\%, with a radial MAE of 0.505\,m/s).
  \end{minipage}
\end{table*}

Table~\ref{tab:supp_runtime} lists the model size and runtime. With the same weights, FP32 inference changes the class macro accuracy on the test set by only 0.1 percentage points (75.6\% compared with 75.5\% for FP16) and leaves the other main metrics unchanged, including the object radial-velocity MAE, geometry mIoU, and ego-speed MAE. It increases the latency by about 55\% (about 76\% for PCG), so we use FP16 (Sec.~\ref{sec:supp_training}). On an RTX 5090, the complete system runs at 12.5\,FPS, faster than the 10\,FPS radar frame rate. We apply no further deployment optimization such as pruning or quantization.

\begin{table}[!t]
  \centering
  \caption{Model size and runtime at batch size 1. Runtime entries are latency (ms) / FPS.}
  \label{tab:supp_runtime}
  \footnotesize
  \setlength{\tabcolsep}{3pt}
  \renewcommand{\arraystretch}{1.08}
  \begin{tabular*}{\columnwidth}{@{\extracolsep{\fill}}lcccc@{}}
    \toprule
    Model & \makecell{Params\\(M)} & \makecell{RTX 3090\\FP16} & \makecell{RTX 3090\\FP32} & \makecell{RTX 5090\\FP16} \\
    \midrule
    PCG & 6.17 & 121.7 / 8.2 & 213.8 / 4.7 & 49.7 / 20.1 \\
    RadarOSV & 18.85 & 174.0 / 5.7 & 270.5 / 3.7 & \textbf{80.2 / 12.5} \\
    \bottomrule
  \end{tabular*}
  \par\smallskip
  \begin{minipage}{\columnwidth}
    \footnotesize
The timing covers network inference only and excludes the signal processing from ADC data to the input cubes. The complete system includes both radar branches and ego-motion estimation with refinement. Peak memory on the RTX 3090 is 5.3\,GB with FP16 and 7.0\,GB with FP32.
  \end{minipage}
\end{table}

\section{Joint Training and Cross-Dataset Transfer}
\label{sec:supp_transfer}

\subsection{Input Frames under Joint Training}
\label{sec:supp_joint}

RaDelft and BRIDGE-Liv use the same cascaded radar model with similar configurations (Table~\ref{tab:supp_datasets}), so the RaDelft data can be processed into RAE cubes with the same pipeline. We therefore train a single PCG on both datasets. With this joint training, three input frames are better than one on both datasets (Table~\ref{tab:supp_frames_joint}).

\begin{table}[!t]
  \centering
  \caption{Number of PCG input frames under joint training (mIoU in \%).}
  \label{tab:supp_frames_joint}
  \footnotesize
  \setlength{\tabcolsep}{3pt}
  \renewcommand{\arraystretch}{1.08}
  \begin{tabular*}{\columnwidth}{@{\extracolsep{\fill}}lcc@{}}
    \toprule
    PCG input & BRIDGE-Liv & RaDelft \\
    \midrule
    1 frame & 29.1 & 25.8 \\
    \textbf{3 frames} & \textbf{29.8} & \textbf{26.5} \\
    \bottomrule
  \end{tabular*}
\end{table}

\subsection{Transfer of the Pretrained PCG to Other Datasets}
\label{sec:supp_pretrain}

We test whether a PCG trained on BRIDGE-Liv transfers to other cascaded radar datasets. We train PCG on RaDelft and on ColoRadar in two ways. The first fine-tunes the PCG pretrained only on BRIDGE-Liv, and the second trains from scratch. Both use the same settings (five epochs, three frames, one power channel, and a learning rate of $2\times10^{-4}$).

Table~\ref{tab:supp_pretrain} reports the results after one, two, and five epochs. Fine-tuning is better than training from scratch in every case. The gain is largest after the first epoch (2.6 points of mIoU on RaDelft and 3.3 on ColoRadar) and remains after five epochs (0.9 and 1.6 points). On RaDelft, one epoch of fine-tuning already gives a higher mIoU than five epochs from scratch (25.4\% compared with 25.0\%). The geometry that PCG learns on BRIDGE-Liv therefore transfers to the cascaded radars of the other two datasets.

\begin{table}[!t]
  \centering
  \caption{Fine-tuning the PCG pretrained on BRIDGE-Liv compared with training from scratch. Entries are mIoU (\%) / 3D Chamfer distance (m).}
  \label{tab:supp_pretrain}
  \footnotesize
  \setlength{\tabcolsep}{2.5pt}
  \renewcommand{\arraystretch}{1.08}
  \begin{tabular*}{\columnwidth}{@{\extracolsep{\fill}}llccc@{}}
    \toprule
    Dataset & Training & Epoch 1 & Epoch 2 & Epoch 5 \\
    \midrule
    RaDelft & Fine-tuned & \textbf{25.4 / 1.37} & \textbf{26.1 / 1.31} & \textbf{25.9 / 1.25} \\
    RaDelft & From scratch & 22.8 / 1.72 & 24.3 / 1.41 & 25.0 / 1.28 \\
    \midrule
    ColoRadar & Fine-tuned & \textbf{28.2 / 1.58} & \textbf{31.5 / 1.43} & \textbf{33.7 / 1.30} \\
    ColoRadar & From scratch & 24.9 / 1.74 & 29.2 / 1.44 & 32.1 / 1.37 \\
    \bottomrule
  \end{tabular*}
\end{table}

\subsection{Zero-Shot Transfer of Class and Velocity to RaDelft}
\label{sec:supp_zeroshot}

RaDelft has no ground truth for the class and velocity of moving objects (Sec.~\ref{sec:supp_datasets}). As a qualitative reference, we apply the trained N2 model directly to the RaDelft test scenes and inspect its class and velocity outputs.

Fig.~\ref{fig:supp_zeroshot} shows six examples, of which (a)--(e) are selected successful cases. We track some objects in the LiDAR point clouds for about 1\,s to check their velocities. The bus in (c) is predicted at 7.3\,m/s and tracked at 7.0\,m/s, the car in (d) at 6.1 and 6.2\,m/s, and the cyclist in (e) at 5.5 and 4.9\,m/s. These examples do not represent the average performance. Pedestrians are often missed, cyclists are often assigned the wrong class, stationary vehicles and roadside structures are sometimes predicted as moving, as in (f), and the velocity errors are larger than on BRIDGE-Liv. N2 already makes the same kinds of errors on BRIDGE-Liv, such as predicting pedestrians as static and underestimating object speed (Fig.~3 of the main paper), and they are more frequent on RaDelft. RaDelft was recorded in a different country, with different road scenes from BRIDGE-Liv. The class and velocity heads trained on BRIDGE-Liv can still be used in part on RaDelft, but reaching their BRIDGE-Liv performance would require fine-tuning with class and velocity annotations of moving objects, which RaDelft does not provide.

\begin{figure}[!t]
  \centering
  \includegraphics[width=\columnwidth]{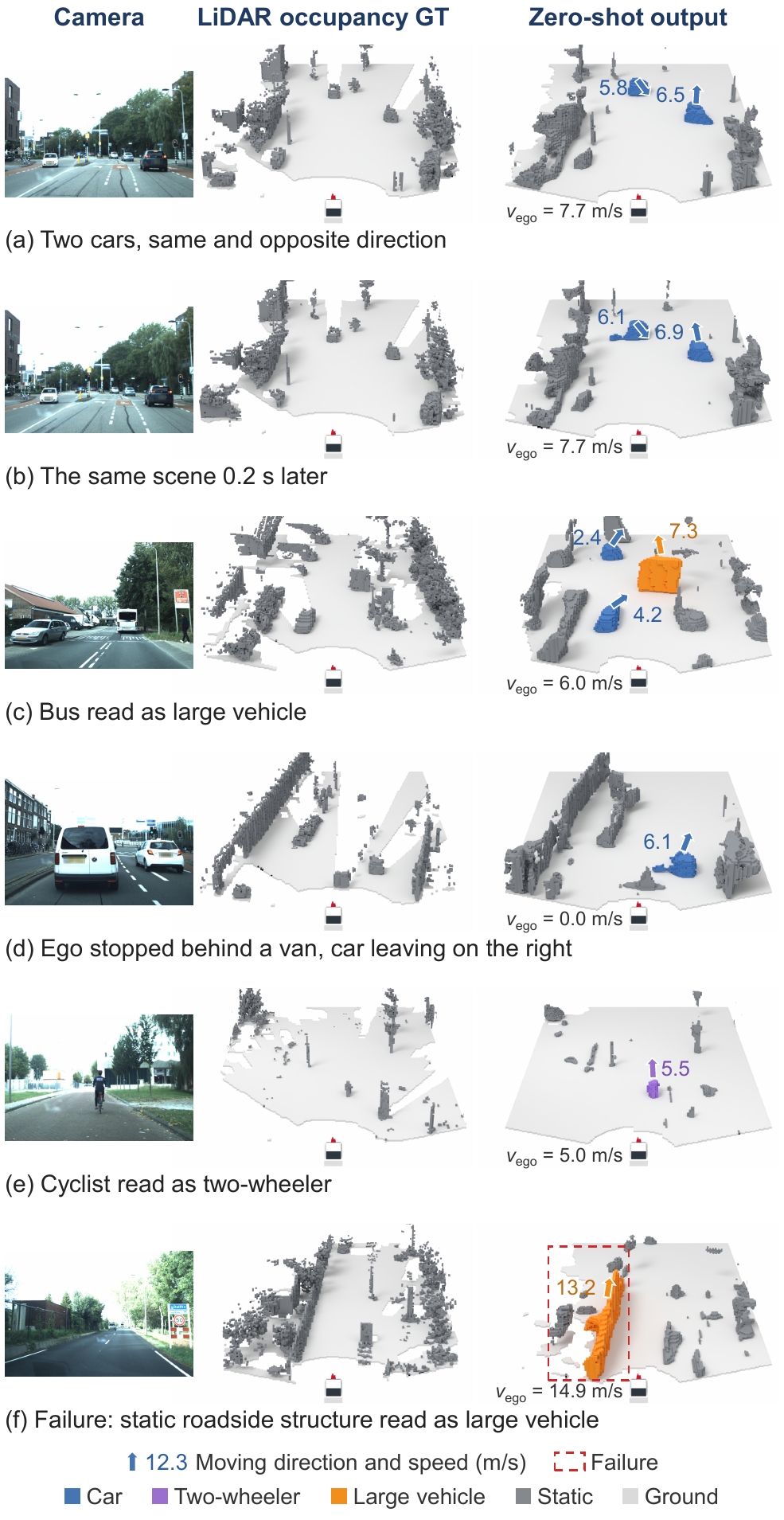}
  \caption{Zero-shot transfer of class and velocity to RaDelft test scenes with N2 (CAS only).
  Each row shows the camera image for reference, the LiDAR occupancy ground truth, and the model output.
  The ground truth has no class labels, so all its objects are shown in gray.
  Arrows and numbers show the predicted velocity over the ground.
  (a, b) Two cars, one moving in the same direction as the ego vehicle and one in the opposite direction, are predicted as car in two frames 0.2\,s apart, with consistent class and velocity.
  (c) A bus is predicted as large vehicle.
  (d) The ego vehicle and the van ahead are stopped. The van stays static, and the car leaving on the right is predicted as car.
  (e) A cyclist is predicted as two-wheeler.
  (f) Failure case. A static roadside structure is predicted as large vehicle.
  Each connected component is shown in a single class.}
  \label{fig:supp_zeroshot}
\end{figure}

\section{Failure Cases}
\label{sec:supp_failure}

Fig.~\ref{fig:supp_failure} shows seven failure cases on BRIDGE-Liv. They fall into three types, and all of them are related to limits of the radar measurement itself.

\noindent\textbf{Slow objects (a--c).} Pedestrians and cars creeping in a queue move at less than 2\,m/s over the ground. Their Doppler is close to that of the static background, and they are predicted as static. This is more common when the ego vehicle is moving.

\noindent\textbf{Neighboring objects (d--f).} When two objects are at the same range and closer in azimuth than the azimuth resolution of SC (8 elements, about 14$^\circ$), their Doppler peaks fall in the same range--azimuth cell and the velocity can be read wrong. In (d), the oncoming car moves at 14.0\,m/s but is read as 5.3\,m/s, close to the 5.5\,m/s of the car beside it. In (f), the speeds of the truck and the car beside it are off by 3 to 4\,m/s. A car right beside a large vehicle can also merge with it into one component, as in (e).

\noindent\textbf{Tangential motion (g).} The tangential velocity has no Doppler measurement. The velocity head regresses it from the fused features (Sec.~\ref{sec:supp_vel_readout}), and the main cue is how the object position changes over the three input frames, which span 0.2\,s. An object with a tangential speed of 2\,m/s moves only 0.4\,m in 0.2\,s, less than the azimuth resolution of CAS at 30\,m (86 elements, about 1.3$^\circ$, or 0.7\,m). In addition, most objects in the data travel along the road and few cross it. The tangential component is therefore less accurate than the radial one, and it is often underestimated at roundabouts and junctions. In (g), the car crossing on the left moves at 5.9\,m/s, of which only 1.5\,m/s is radial, and its speed is read as 1.2\,m/s. The other two cars move slowly (2.4 and 1.5\,m/s). Their radial components are close to the ground truth, but their tangential components are wrong, and the arrows are off by more than 45$^\circ$. A longer input window may improve the tangential estimate.

Outside the red boxes, the other moving objects in Fig.~\ref{fig:supp_failure} have classes and velocities close to the ground truth, and no static structure is predicted as a moving class. Over the whole test set, 91.3\% of the 13{,}306 annotated moving objects (Sec.~\ref{sec:supp_eval}) are predicted as a moving class, and 99.2\% of these have the correct class.

\begin{figure}[!t]
  \centering
  \includegraphics[width=\columnwidth]{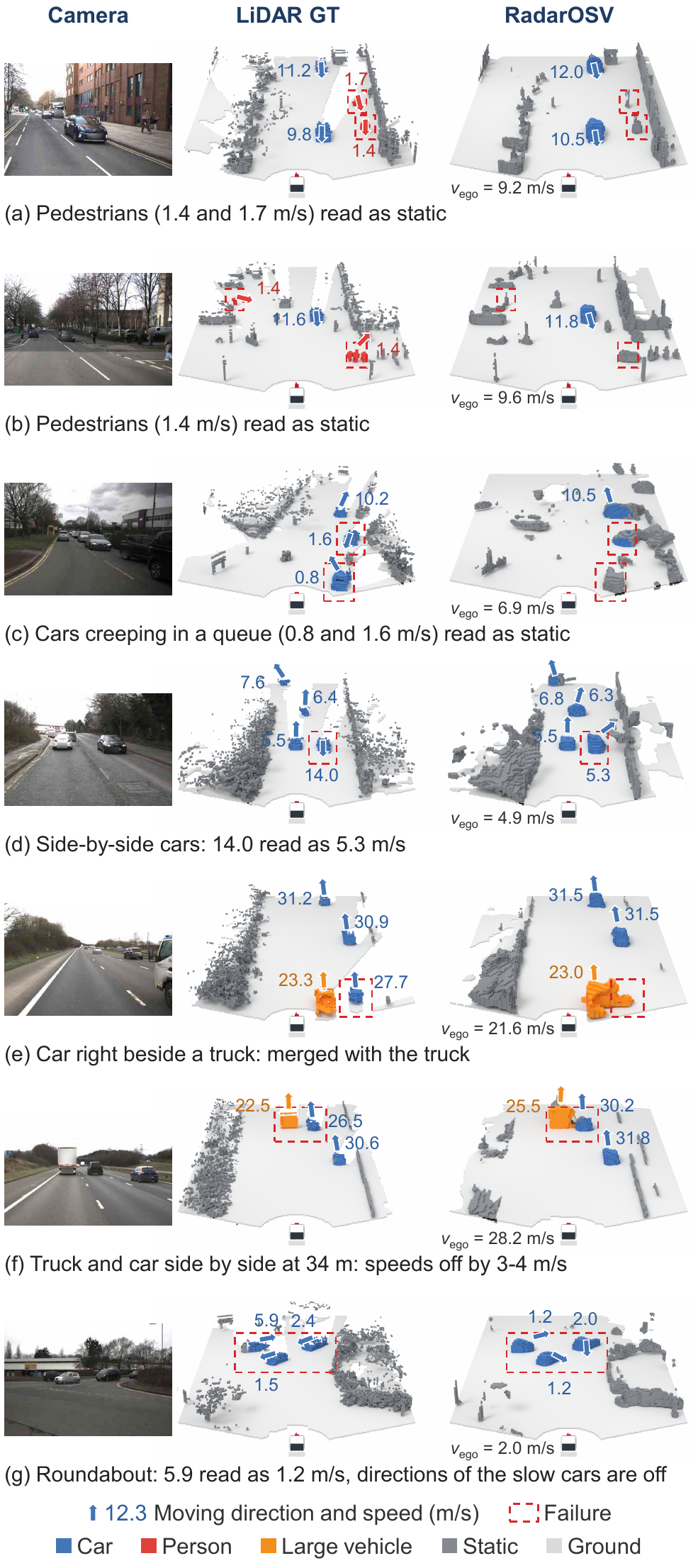}
  \caption{Failure cases on BRIDGE-Liv.
  The camera images are for reference, and red dashed boxes mark the failed objects.
  (a--c) Slow objects. (d--f) Neighboring objects. (g) Tangential motion.
  Arrows and numbers show the velocity of each object, taken as the median over its voxels (Sec.~\ref{sec:supp_eval}). An object predicted as static has no arrow.
  For display, a connected component is drawn in a single moving class when that class makes up most of it, a moving class that makes up a very small part is not drawn, and all other voxels are drawn as predicted.}
  \label{fig:supp_failure}
\end{figure}

\section{Range-Binned Geometry Results}
\label{sec:supp_range}

Table~\ref{tab:supp_range} reports the geometry metrics of RadarOSV on BRIDGE-Liv in three range bins. The mIoU drops from 46.0\% within 20\,m to 12.7\% beyond 40\,m. At longer range, the returns are weaker, the same angular resolution covers a larger width, and the LiDAR labels themselves are sparser. Over the three bins together, the mIoU is 30.7\%, which is the RadarOSV row in Table~4 of the main paper.

\begin{table}[!t]
  \centering
  \caption{Geometry metrics of RadarOSV on BRIDGE-Liv by range (in \%). The range is the horizontal distance from the voxel center to CAS, and the last bin extends to the maximum radar range of about 49\,m.}
  \label{tab:supp_range}
  \footnotesize
  \setlength{\tabcolsep}{3pt}
  \renewcommand{\arraystretch}{1.08}
  \begin{tabular*}{\columnwidth}{@{\extracolsep{\fill}}lccc@{}}
    \toprule
    Range & mIoU & Ground IoU & Object IoU \\
    \midrule
    0--20\,m & 46.0 & 65.9 & 26.0 \\
    20--40\,m & 26.2 & 37.5 & 15.0 \\
    $\geq$40\,m & 12.7 & 19.3 & 6.1 \\
    \bottomrule
  \end{tabular*}
\end{table}

\end{document}